\pdfoutput=1

\documentclass[11pt]{article}

\usepackage[]{acl}

\usepackage{times}
\usepackage{latexsym}

\usepackage[T1]{fontenc}

\usepackage[utf8]{inputenc}

\usepackage{microtype}

\usepackage{graphicx}
\usepackage{caption}
\usepackage{subcaption}
\usepackage{amsmath}
\usepackage{amsthm}
\usepackage{amssymb}
\usepackage{resizegather}
\usepackage{multicol}
\usepackage{booktabs}
\usepackage{adjustbox}

\title{
Re-Examining System-Level Correlations of Automatic Summarization Evaluation Metrics}

\author{Daniel Deutsch, Rotem Dror, and Dan Roth \\
        Department of Computer and Information Science \\ 
        University of Pennsylvania \\
        \texttt{\{ddeutsch,rtmdrr,danroth\}@seas.upenn.edu}}

\newif\ifcomments
\commentstrue
\ifcomments
    \providecommand\dd[1]{\textcolor{blue}{[DD: #1]}}
    \providecommand\dr[1]{\textcolor{blue}{[DR: #1]}}
    \providecommand\rd[1]{\textcolor{purple}{[RD: #1]}}
    \providecommand\sihao[1]{\textcolor{magenta}{[SC: #1]}}
    \providecommand\todo[1]{\textcolor{red}{[TODO: #1]}}
\else
    \providecommand{\dd}[1]{}
    \providecommand{\dr}[1]{}
    \providecommand{\rd}[1]{}
    \providecommand\sihao[1]{}
    \providecommand{\todo}[1]{}
\fi

\newcommand{\qaeval}{QA\-Eval}
\newcommand{\bertscore}{BERT\-Score}
\newcommand{\summeval}{Summ\-Eval}
\newcommand{\realsumm}{REAL\-Summ}
\newcommand{\rsys}{r_\textsc{sys}}
\newcommand{\mtest}{M_\textrm{test}}
\newcommand{\mjudged}{M_\textrm{jud}}

\begin{document}

\maketitle

\begin{abstract}
How reliably an automatic summarization evaluation metric replicates human judgments of summary quality is quantified by system-level correlations.
We identify two ways in which the definition of the system-level correlation is inconsistent with how metrics are used to evaluate systems in practice and propose changes to rectify this disconnect.
First, we calculate the system score for an automatic metric using the full test set instead of the subset of summaries judged by humans, which is currently standard practice.
We demonstrate how this small change leads to more precise estimates of system-level correlations.
Second, we propose to calculate correlations only on pairs of systems that are separated by small differences in automatic scores which are commonly observed in practice.
This allows us to demonstrate that our best estimate of the correlation of ROUGE to human judgments is near 0 in realistic scenarios.
The results from the analyses point to the need to collect more high-quality human judgments and to improve automatic metrics when differences in system scores are small.\footnote{
   Our code is available at \url{http://cogcomp.org/page/publication_view/973}.
}
\end{abstract}

\section{Introduction}
\label{sec:intro}

Automatic evaluation metrics are the most common method that researchers use to quickly and cheaply approximate how humans would rate the quality of a summarization system \citep[][among others]{Lin04,LouisNe13,ZPLGME19,ZKWWA20,DeutschBeRo21}.
The quality of a metric --- how similarly it replicates human judgments of systems --- is quantified by calculating the correlation between the metric's scores and human judgments on a set of systems, known as the system-level correlation \citep{LouisNe13,DeutschDrRo21}.

Accurately estimating system-level correlations is critically important.
Summarization researchers use automatic metrics during system development to make decisions about which ideas work and which do not, and systems from different research groups are ranked by automatic metrics to define which system is the ``state-of-the-art.''
If we do not have precise estimates of metric quality, it is not clear how much trust the community should put in such evaluation methodologies. 

At present, there are disconnects between how automatic metrics are evaluated and how they are used to evaluate systems.
First, the metrics' scores which are used in practice are not the ones which are evaluated in system-level correlations;
researchers compare systems based on metric scores calculated on the entire test set but calculate scores for system-level correlations when evaluating metrics on a much smaller subset of judged summaries.
Second, metrics are evaluated in a setting that is much easier than how they are actually used.
Metric correlations are calculated using systems that vary greatly in quality, whereas researchers compare new systems to recent work, which are likely to be very close in quality.
Discriminating between two systems of similar quality is much harder than doing so between low and high quality systems.

In this work, we re-examine how system-level correlations are calculated and propose two independent changes to make the evaluation of metrics better aligned to how they are actually used to evaluate systems.

First, we propose to modify the system-level correlation definition to use the entire test set to calculate the system scores for automatic metrics instead of only the subset of summaries judged by humans (\S\ref{sec:corr}).
With this change, the scores which are used to compare systems are directly evaluated, and we further demonstrate how the precision of our estimate of system-level correlations improves as a result.
Calculating system scores over a larger number of instances reduces the variance of the scores, which results in confidence intervals (CIs) for the correlations that are 16-51\% more narrow on average (\S\ref{sec:confidence_intervals}).

Second, we redefine a high quality metric to be one for which a small difference in score reliably indicates a difference in quality (\S\ref{sec:delta_correlations}).
Then, instead of calculating the correlation with all available system pairs, we only evaluate with pairs of systems whose automatic metric scores differ by some threshold. 
This allows us to show, for example, that a ROUGE-1 score difference of less than 0.5 between systems has almost no correlation to how humans would rank the same two systems according to our best estimates (\S\ref{sec:delta_correlations_subsec}).
For two other metrics, \bertscore{} \citep{ZKWWA20} and \qaeval{} \citep{DeutschBeRo21}, we show their correlations calculated on system pairs of similar quality are much worse than under the standard correlation definition.
These results cast doubt on how reliable automatic evaluation metrics are for measuring summarization system quality in realistic scenarios.

Our analyses point to the need to collect more high-quality human judgments of summaries in order to have more accurate estimates of metric correlations as well as the need to improve the ability of automatic metrics to discriminate between similarly performing systems.

\section{Background}
\label{sec:background}

Automatic evaluation metrics are most commonly used to argue that one summarization system is better than another, typically by showing that the value of a metric improves with the ``better'' system.
How similarly automatic metrics replicate human judgments of system quality is quantified by system-level correlations as follows.

\begin{figure}
    \centering
    \includegraphics[width=\columnwidth]{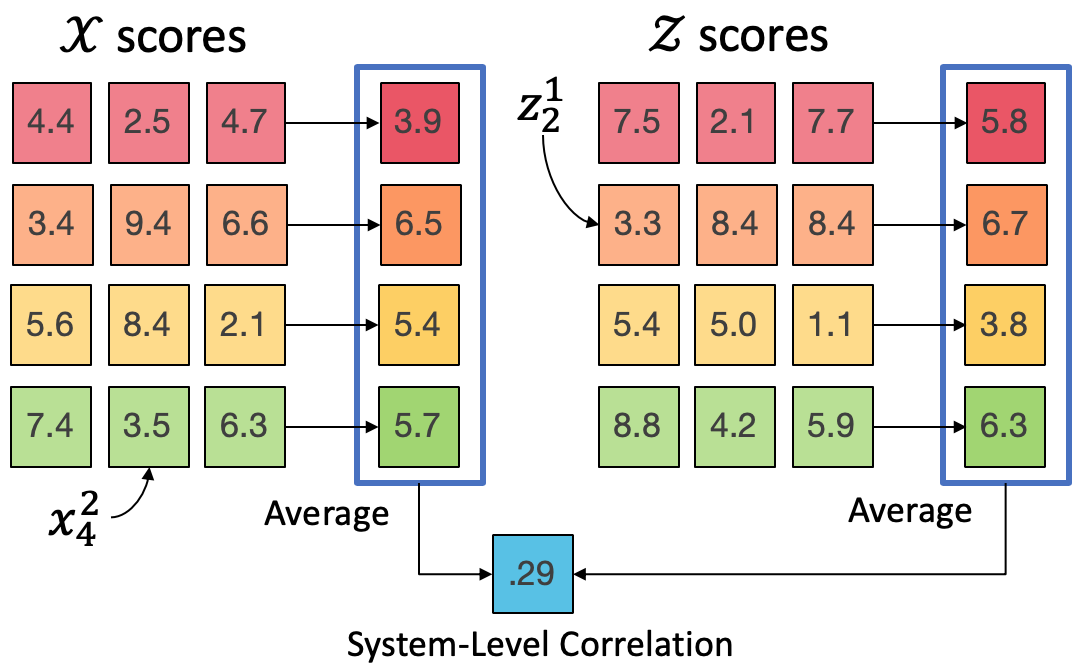}
    \caption{The system-level correlation is calculated between the average $\mathcal{X}$ and $\mathcal{Z}$ scores on a set of summarization systems.
    $x_i^j$ and $z_i^j$ are the scores for the summary produced by system $i$ (represented by rows) on input document $j$ (represented by columns).}
    \label{fig:corr_diagram}
\end{figure}

The summaries from $N$ systems on $\mjudged$ input documents are judged by humans $\mathcal{Z}$ and scored with an automatic metric $\mathcal{X}$.
Then, the system-level correlation between $\mathcal{X}$ and $\mathcal{Z}$ is calculated as
\begin{gather*}
    \rsys = \textsc{Corr}\left(\left\{\left(\frac{1}{\mjudged}\sum_j^{\mjudged} x^j_i, \frac{1}{\mjudged}\sum_j^{\mjudged} z^j_i\right)\right\}_{i=1}^N\right)
\end{gather*}
where $x_i^j$ and $z_i^j$ are the scores of $\mathcal{X}$ and $\mathcal{Z}$ for the summary produced by the $i$-th system on the $j$-th input document and $\textsc{Corr}$ is some correlation function.
See Fig.~\ref{fig:corr_diagram} for an illustration of this calculation.

In this work, we use Kendall's $\tau$ (the ``b'' variant\footnote{\url{https://docs.scipy.org/doc/scipy/reference/generated/scipy.stats.kendalltau.html}}) as the correlation function because we are most concerned with a metric's ability to correctly determine whether one system is better than another since that is how metrics are used in practice.
Kendall's $\tau$ is computed based on the number of system pairs out of ${N \choose 2}$ which are ranked the same by $\mathcal{X}$ and $\mathcal{Z}$.
It is defined as
\begin{equation}
    \tau = \frac{P - Q}{\sqrt{(P + Q + T) \cdot (P + Q + U)}}
\end{equation}
where $P$ and $Q$ are the number of pairs ranked the same or different by $\mathcal{X}$ and $\mathcal{Z}$, respectively, and $T$ and $U$ are the number of ties only in $\mathcal{X}$ or $\mathcal{Z}$, respectively.

Because the computation of $\rsys$ involves randomness --- its value depends on which $\mjudged$ input documents (and even which $N$ systems) were used --- it is only an approximation of the true correlation between $\mathcal{X}$ and $\mathcal{Z}$.
As such, \citet{DeutschDrRo21} proposed various methods for calculating confidence intervals for $\rsys$.
For instance, their \textsc{Boot-Inputs} method uses bootstrapping to repeatedly resample the $\mjudged$ input documents used to calculate $\rsys$, thereby calculating a confidence interval for the true $\rsys$ value for $\mathcal{X}$ and $\mathcal{Z}$.

\paragraph{Datasets}
The datasets that are used in this paper's analyses are \summeval{} \citep{FKMSR21} and \realsumm{} \citep{BGALN20}, two recently collected datasets with human annotations for summary quality collected from the CNN/DailyMail dataset \citep{NallapatiBoCiCaBi16}.
\summeval{} has $\mjudged = 100$ summaries annotated with a summary relevance score for $N = 16$ systems.
\realsumm{} has $\mjudged = 100$ summaries annotated with a Lightweight Pyramid score \citep{ShapiraGaGaRoPaBaAmDa19} for $N = 25$ systems.
We correlate the scores of the automatic metrics to these annotations.
The CNN/DailyMail test split has $11,490$ instances.

\paragraph{Automatic Metrics}
Our experiments will analyze three different reference-based automatic evaluation metrics which were chosen because they were demonstrated to have the best correlations with human judgments on the \summeval{} and \realsumm{} datasets \citep{DeutschDrRo21}.
ROUGE-$n$ \citep{Lin04} evaluates a generated summary by calculating an F$_1$ score on the number of $n$-grams it has in common with a human-written reference summary.
BERTScore \citep{ZKWWA20} aligns the generated and reference summaries' tokens based on their BERT embeddings \citep{DCLT19} and calculates a score based on the similarity of the aligned tokens' embeddings.
\qaeval{} \citep{DeutschBeRo21} compares the two summaries by automatically generating questions from the reference and calculating what proportion of those questions are answered correctly by the generated summary.
\section{Evaluating with All Available Instances}
\label{sec:corr}

Although the above definition of the system-level correlation has been used by recent meta-evaluation studies of metrics \citep{BGALN20,FKMSR21,DeutschDrRo21}, there is a disconnect between how the automatic metrics are evaluated and how they are used in practice.

Researchers who develop summarization systems evaluate those systems with automatic metrics on all $\mtest$ test instances, not just the subset of $\mjudged$ instances which were judged by humans. 
Evaluating a system on a larger number of summaries may end up changing the system's score, which could potentially alter the overall ranking of a set of systems.
Therefore, the rankings that are used by practitioners to determine system quality are not the ones which are being evaluated in the standard definition of system-level correlation.\footnote{
    We suspect this methodology is an artifact of how system-level correlations were first calculated for summarization in the DUC shared tasks when the dataset sizes were small enough that $\mjudged = \mtest$ \citep[e.g.,][]{DangOw08}.
} 

To that end, we propose to modify the correlation definition to use all $\mtest$ instances to calculate the system scores for the automatic metrics.
That is (differences in bold):
\begin{gather*}
    r_\textsc{sys} = \textsc{Corr}\left(\left\{\left(\frac{1}{\mathbf{M}_\textrm{\bf{test}}}\sum_j^{\mathbf{M}_\textrm{\bf{test}}} x^j_i, \frac{1}{\mjudged}\sum_j^{\mjudged} z^j_i\right)\right\}_{i=1}^N\right)
\end{gather*}
In practice with modern, large-scale datasets, this minor change could mean estimating system quality based on $\approx$10k inputs instead of around 100.
This new definition now properly evaluates the way metrics are actually used by researchers.

We expect that scoring systems with $\mtest$ inputs instead of $\mjudged$ should lead to a better estimate of the true automatic metric score, which would in turn result in a lower-variance estimate of the correlation between $\mathcal{X}$ and $\mathcal{Z}$ in the form of smaller confidence intervals for $\rsys$.
In the next sections, we carry out analyses to demonstrate that this is true.

\begin{figure}
    \centering
    \includegraphics[width=\columnwidth]{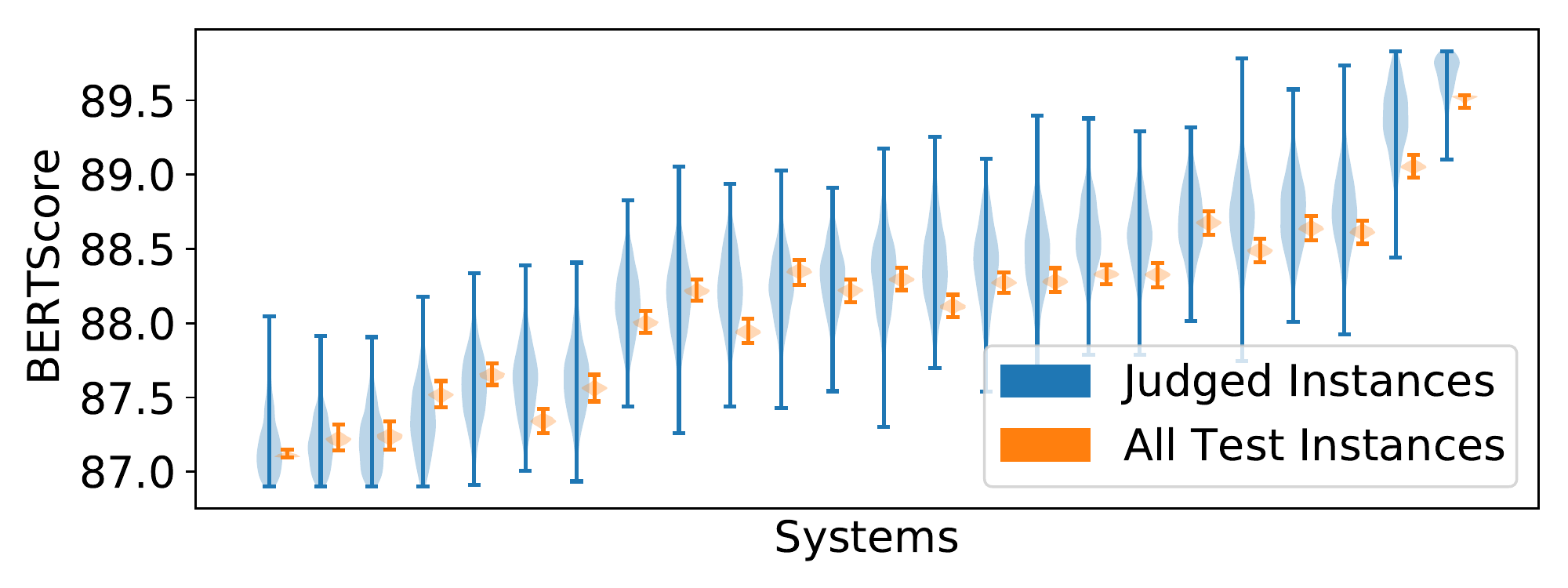}
    \caption{The bootstrapped 95\% confidence intervals for the BERT\-Score of each system in the \realsumm{} dataset using $\mjudged$ judged instances in blue and $\mtest$ instances in orange.
    Evaluating systems with $\mtest$ instances leads to far better estimate of their true scores.
    }
    \label{fig:score_variance}
\end{figure}

\begin{figure*}
\centering
\begin{subfigure}{.48\textwidth}
  \centering
  \includegraphics[width=\textwidth]{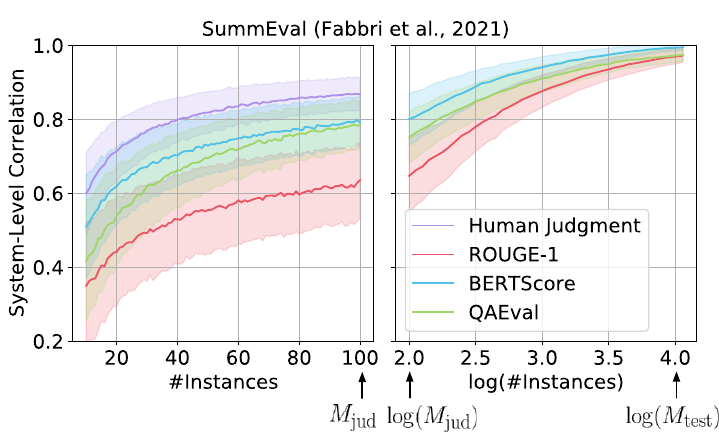}
  \label{fig:sub1}
\end{subfigure}
\hspace{2.5mm}
\begin{subfigure}{.48\textwidth}
  \centering
  \includegraphics[width=\textwidth]{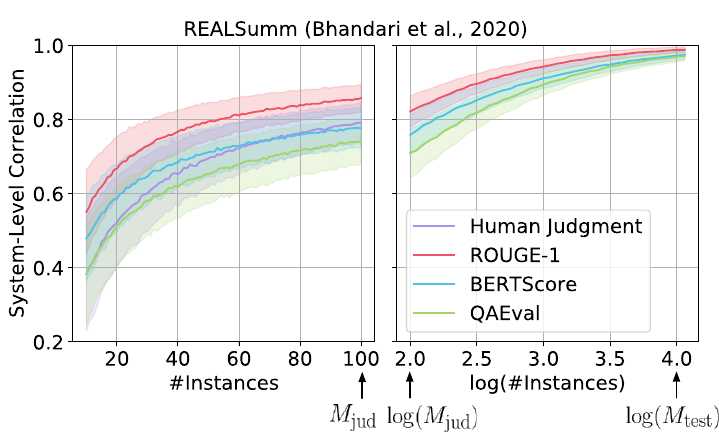}
  \label{fig:sub2}
\end{subfigure}
\caption{
Bootstrapped estimates of the stabilities of the system rankings for automatic metrics and human annotations on \summeval{} (left) and \realsumm{} (right).
The $\tau$ value quantifies how similar two system rankings would be if they were computed with two random sets of $M$ input documents.
When all $\mtest$ test instances are used, the automatic metrics' rankings become near constant.
The error regions represent $\pm 1$ standard deviation.
}
\label{fig:ranking_variance}
\end{figure*}

\subsection{Reducing Automatic Metric Variance}
First, we empirically show that scoring systems with $\mtest$ instances instead of $\mjudged$ does indeed reduce the variance of the estimate of the automatic metric scores and subsequently increases the stabilities of the system rankings.

Ideally, the $\mathcal{X}$ score for a system would be its ``oracle'' $\mathcal{X}$ score, equal to the expected value of $\mathcal{X}$ for a document sampled from the latent distribution over documents defined by the dataset (e.g., a system's ROUGE score on an infinite number of examples from a dataset).
Since this cannot be calculated, it is approximated by averaging the $\mathcal{X}$ score on a sample (i.e., either the $\mjudged$ or $\mtest$ input documents).
Because $\mtest \gg \mjudged$, we expect that the variance of this estimate using $\mtest$ inputs should be lower than when using $\mjudged$.

To quantify this, we calculated the variance of estimating the oracle $\mathcal{X}$ score using both $\mjudged$ and $\mtest$ input documents via bootstrapping.
We randomly sampled $M$ input documents with replacement, recomputed the system scores, and calculated the variance of those scores over 1k iterations.
For all three metrics on both datasets, we found around a 99\% reduction in the variance when $\mtest$ inputs were used instead of $\mjudged$, clearly demonstrating that evaluating systems with $\mtest$ inputs results in a better estimate of the system scores.
In Fig.~\ref{fig:score_variance}, this is visualized for \bertscore{} on the \realsumm{} dataset.

However, because we are interested in evaluating the metrics' rankings, we also quantify how much of an effect this reduction in variance has on the stability of the system rankings induced by $\mathcal{X}$.
Similarly to the system scores, there is an oracle ranking of systems for $\mathcal{X}$, equal to the ordering of systems by their respective oracle $\mathcal{X}$ scores (e.g., systems sorted by their ROUGE scores calculated on an infinite number of examples from a dataset).
As the variance of the system score estimates decreases, the computed ranking of systems should begin to converge to the oracle $\mathcal{X}$ ranking.
We aim to understand to what extent this happens if $\mtest$ instances are used for evaluation instead of $\mjudged$.

To quantify this notion, we calculate the Kendall's $\tau$ between two system rankings for $\mathcal{X}$ that were based on two sets of $M$ input documents, each sampled with replacement from the set of available documents.
This simulates how much the system rankings would change if the evaluation procedure was run twice, each time with $M$ random input documents.
This quantity is calculated 1k times for various values of $M$ and plotted in Fig.~\ref{fig:ranking_variance}.

As $M$ approaches $\mtest$, the automatic metrics' $\tau$ values approach 1, which is much higher than the respective values at $\mjudged$, typically around 0.6-0.8.
A value near 1 means that the rankings calculated using $\mtest$ inputs are almost constant, implying the rankings have converged to the oracle ranking.
Therefore, the reduction in variance from evaluating on $\mtest$ instances does indeed greatly stabilize the system rankings.

Fig.~\ref{fig:ranking_variance} also contains the same analysis performed for the human judgments $\mathcal{Z}$ in both datasets, although it is limited to a maximum of $\mjudged$ input documents.
We see that on both datasets the judgments' rankings are still quite variable, reaching a maximum $\tau$ of around 0.8-0.85.

\subsection{Confidence Interval Analysis}
\label{sec:confidence_intervals}
Next, we show that the improved estimate of system scores leads to a more precise estimate of $\rsys$ by demonstrating the widths of the confidence intervals for $\rsys$ decrease.

The confidence intervals for $\rsys$ calculated using bootstrapping methods proposed by \citet{DeutschDrRo21} are rather wide.
For instance, the 95\% CI for ROUGE-2 on \summeval{} is $[-.09,.84]$, demonstrating a rather high level of uncertainty in its value.
This is problematic because it means we do not have a good picture of how reliable automatic evaluation metrics are.
Reducing the width of the CIs will help us better understand the true metric quality.

We suspect that the large width of the confidence interval is due to the variance of the system rankings of the automatic metrics and human judgments.
The more unstable the rankings are with respect to the $M$ inputs, the larger the variance of the estimate of $\rsys$ should be since very different system rankings would be compared on each bootstrapping iteration.
\citet{DeutschDrRo21} used $\mjudged$ input documents to calculate their CIs.
Therefore, we expect the improved stability of the automatic metric system rankings from evaluating on $\mtest$ instances should result in a more narrow confidence interval for $\rsys$ since some noise has been removed from this computation.

\begin{figure}
    \centering
    \includegraphics[width=\columnwidth]{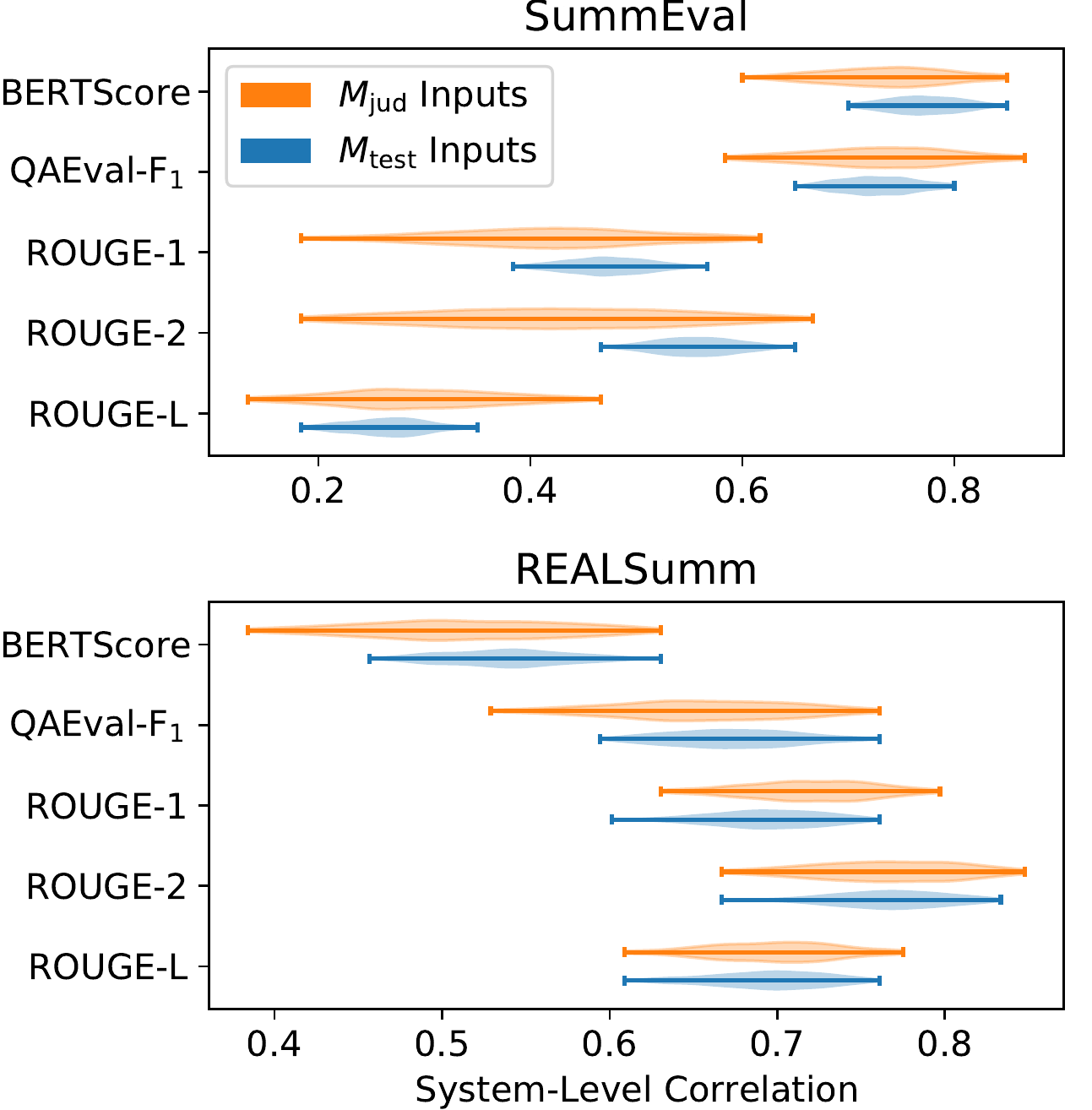}
    \caption{95\% confidence intervals for $\rsys$ calculated with the \textsc{Boot-Inputs} resampling method when the system rankings for the automatic metrics are calculated using only the judged data (orange) versus the entire test set (blue).
    Scoring systems with more summaries leads to better (more narrow) estimates of $\rsys$.}
    \label{fig:ci_boot_input}
\end{figure}

To demonstrate this, we calculated 95\% CIs for $\rsys$ using the \textsc{Boot-Input} method on \summeval{} and \realsumm{} using both $\mjudged$ and $\mtest$ input documents, shown in Fig.~\ref{fig:ci_boot_input}.
We find that the widths of the CIs shrank on average by 51\% on \summeval{} and 16\% on \realsumm{}.
The largest decrease in width is in the ROUGE family of metrics on \summeval{}, potentially because that metric and dataset combination saw the biggest improvement in ranking stability (see Fig.~\ref{fig:ranking_variance}).
Thus, the improved estimate of the system scores did result in more precise estimates of $\rsys$.
We repeated this analysis using the other bootstrapping methods proposed by \citet{DeutschDrRo21}, and the results are discussed in Appendix~\ref{sec:appendix_cis}.

\subsection{Conclusions \& Recommendations}
\label{sec:recs_1}
By estimating system quality using automatic metrics on all available instances instead of only those which were judged, we showed that the variances of the system scores and subsequent rankings reduce significantly, resulting in better estimates of $\rsys$.
Because this methodology additionally directly evaluates the system scores used by researchers, we recommend future work do the same.

In order to continue to improve the estimate of $\rsys$, as much variance as possible needs to be removed from the system rankings.
Evaluating systems using $\mtest$ instances removed a large amount of variance from the automatic metric rankings, but as demonstrated in Fig.~\ref{fig:ranking_variance}, the human judgments still have a large amount of variance.

The human rankings' variances can either be reduced by judging more summaries per system or making the judgments more consistent.
Since the human rankings' stabilities in Fig.~\ref{fig:ranking_variance} are mostly beginning to plateau --- especially for \summeval{} --- it may be prohibitively expensive to collect a sufficient number of judgments to better stabilize the rankings \citep{WeiJi21}.
Therefore, we expect the more feasible solution is to improve the consistency of the human judgments, for example by better training the annotators or improving the annotation protocol.

\section{Evaluating with Realistic System Pairs}
\label{sec:delta_correlations}
Next, we argue that the set of systems used to evaluate metrics is not reflective of how metrics are used in practice and propose a new system-level correlation variant to address this problem.

\subsection{Evaluating with All System Pairs}

\begin{figure}
    \centering
    \includegraphics[width=\columnwidth]{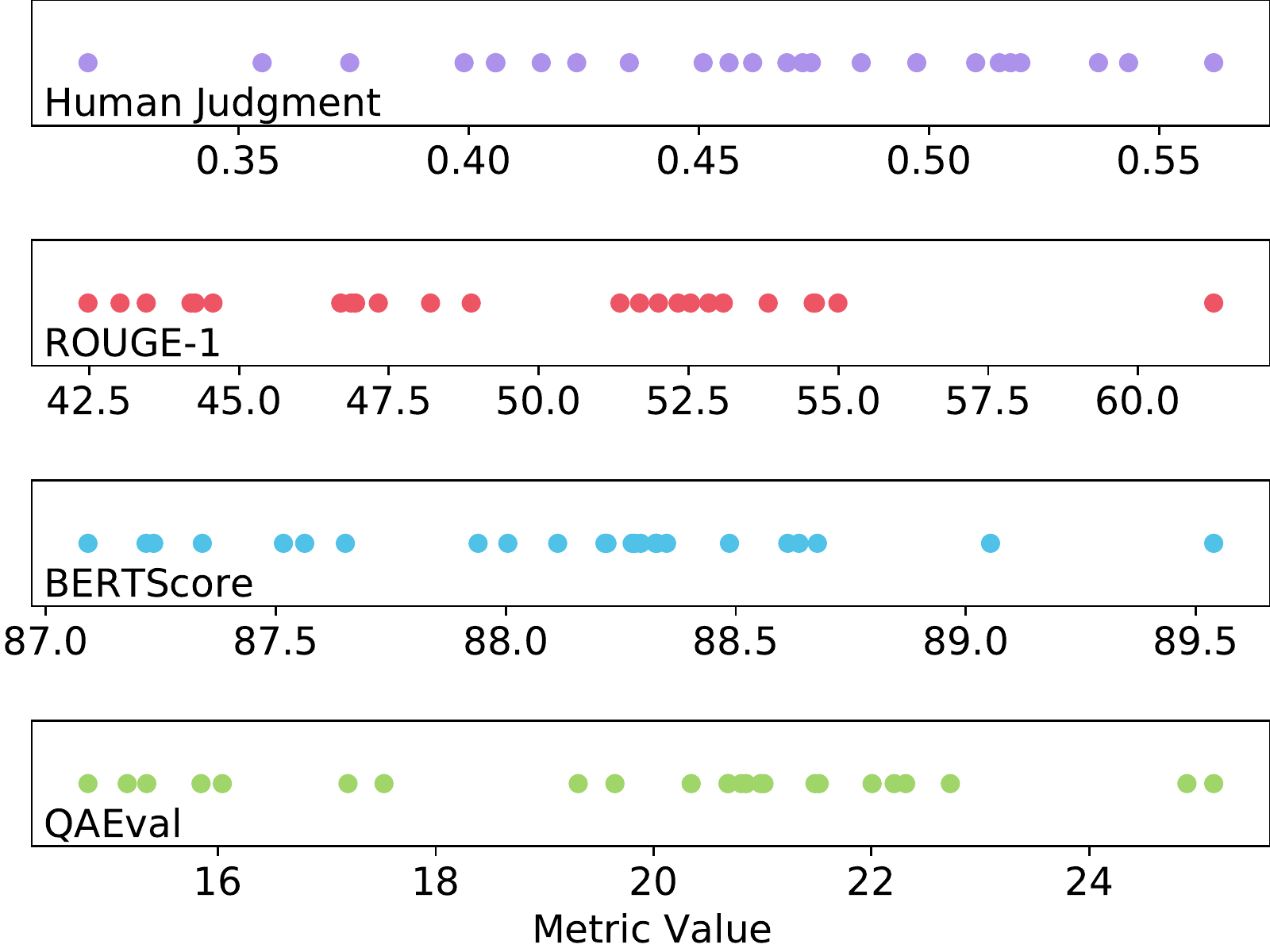}
    \caption{
    The systems (each represented by a point) on the two datasets (shown here for \realsumm{}) are rather diverse in quality as measured by both human judgments and automatic metrics.
    }
    \label{fig:score_distributions}
\end{figure}

The $N$ systems which are used for calculating system-level correlations are typically those which participated in a shared task, as in DUC/TAC \citep[][among others]{DangOw08}, or those which have been published in the previous 3-4 years \citep{BGALN20,FKMSR21}.
As such, they are typically rather diverse in terms of their qualities, both as rated by human annotators and automatic metrics.

The system scores of all of the systems in the \realsumm{} dataset as evaluated by humans and automatic metrics are shown in Fig.~\ref{fig:score_distributions}.
Clearly, the scores are rather diverse.
For example, the systems cluster into low, medium, and high quality groups (with an additional outlier) as evaluated by ROUGE.
A difference of around 5 ROUGE points between them is a rather large gap for ROUGE scores.

The standard definition for a high quality evaluation metric is one which correctly ranks a set of systems with respect to human judgments.
As such, the implementation of the system-level correlation calculated with Kendall's $\tau$ will rank all $N$ systems according to the human judgments and an automatic metric, then count how many pairs were ranked the same out of all {$N \choose 2$} pairs (see \S\ref{sec:background}).
As a consequence, even pairs of systems which are separated by a large margin according to the automatic metric --- likely systems with a clear difference in quality --- are included in the evaluation.
Therefore, automatic metrics are rewarded for correctly ranking such ``easy'' system pairs.

\subsection{Evaluating with Realistic Pairs}
\label{sec:delta_correlations_subsec}

\begin{figure*}[t]
    \centering
    \includegraphics[width=\textwidth]{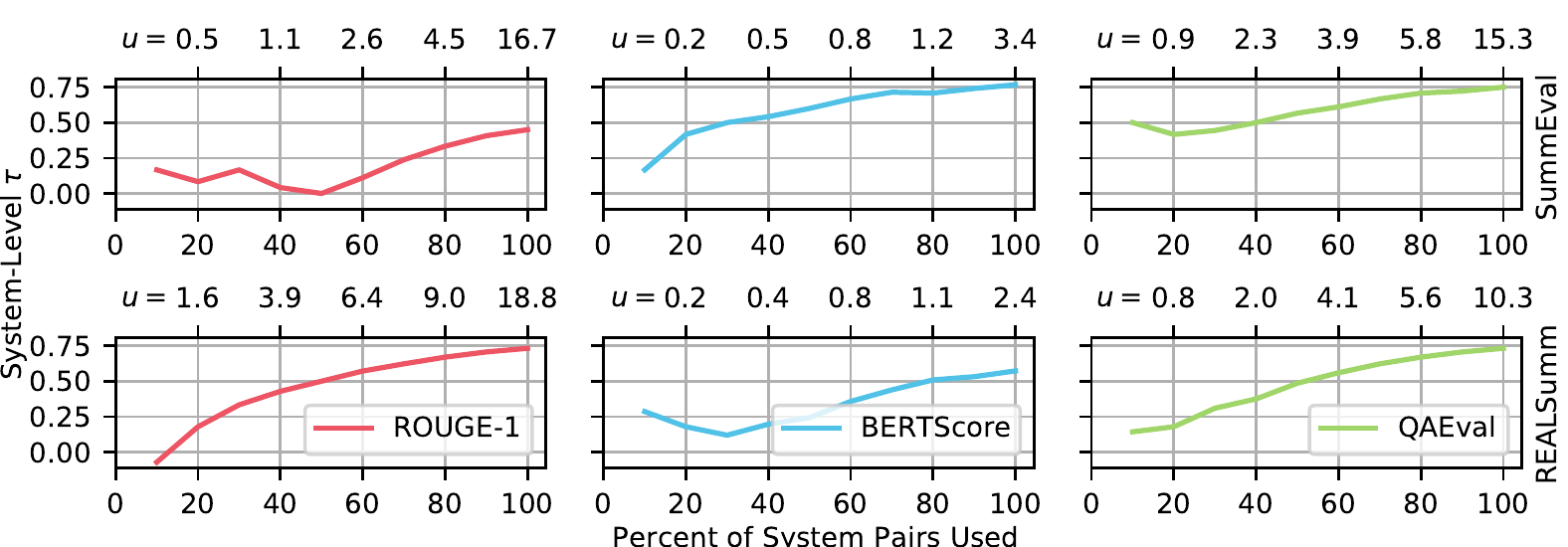}
    \caption{
    The $\rsys\Delta(\ell,u)$ correlations on the \summeval{} (top) and \realsumm{} (bottom) datasets for $\ell = 0$ and various values of $u$ (additional combinations of $\ell$ and $u$ can be found in Appendix~\ref{sec:appendix_heatmaps}).
    The $u$ values were chosen to select the $10\%, 20\%, \dots, 100\%$ of the pairs of systems closest in score.
    Each $u$ is displayed on the top of each plot.
    For instance, 20\% of the ${N \choose 2}$ system pairs on \summeval{} are separated by $<0.5$ ROUGE-1, and the system-level correlation on those pairs is around 0.08.
    As more systems are used in the correlation calculation, the allowable gap in scores between system pairs increases, and are therefore likely easier to rank, resulting in higher correlations.
    }
    \label{fig:delta_plots}
\end{figure*}

This standard evaluation setting does not reflect how summarization metrics are actually used by researchers.
New systems are typically only slightly better than previous work.
Based on a survey of recent summarization papers in *ACL conferences (see Appendix~\ref{sec:survey}), we found that the average improvement over baseline/state-of-the-art models that was reported on the CNN/Dailymail dataset was on average 0.5 ROUGE-1.
It is rarely the case that the improvement in automatic metrics is very large.
Therefore, evaluating metrics using pairs of systems which \emph{are} separated by a large margin does not reflect the reality that metrics are very frequently used to compare those separated by a small margin.
Including ``easy'' system pairs in the system-level correlation likely overestimates the quality of the metrics in settings which occur in practice.

To that end, we redefine a high quality evaluation metric to be one for which a small difference in scores reliably indicates a difference in quality.
We quantify this by proposing a variant of the system-level $\tau$ which is calculated between system pairs which are separated by a pre-defined automatic metric score margin.
Instead of using all {$N \choose 2$} system pairs, only pairs whose difference in scores falls within the margin are used to calculate the system-level correlation.
We denote this correlation variant as $\rsys\Delta(\ell,u)$ where $\ell$ and $u$ are the lower- and upper-bounds of the allowable differences in automatic metrics' scores.
This would enable, for example, evaluating how well ROUGE correlates to human judgments on system pairs that are separated by 0.0-0.5 ROUGE points, thereby directly evaluating the scenario in which ROUGE is used to make decisions about system quality.

In Fig.~\ref{fig:delta_plots} we report the $\rsys\Delta(\ell,u)$ correlations for $\ell = 0.0$ and various values of $u$ on both the \summeval{} and \realsumm{} datasets (more combinations of $\ell$ and $u$ are included in Appendix~\ref{sec:appendix_heatmaps}).
That is, we evaluate $\rsys$ only on system pairs which are separated by at most an automatic score of $u$.
The values of $u$ were selected by picking the minimum $u$ which would result in evaluating on $10\%, 20\%, \dots, 100\%$ of the ${N \choose 2}$ possible system pairs closest in score to be consistent across all three metrics.

The correlations for each metric on the system pairs closest in score are far lower than the correlations evaluated on all of the system pairs.
For instance, the correlation of \bertscore{} on \summeval{} with the closest 20\% of system pairs ($u \approx 0.2$) is only 0.42 compared to 0.77 under the standard definition of $\rsys$.
Thus, it is clear that the metrics are much less reliable approximations of human judgments when the system scores are close than was previously known.
Evaluating on all possible system pairs leads to an overly optimistic view of automatic metric quality.

The $\rsys\Delta(\ell,u)$ correlation of ROUGE for $\ell=0.0$ and $u=0.5$ --- a typical improvement reported by researchers --- is $0.08$ and $0.0$ on the \summeval{} and \realsumm{} datasets.
Therefore, these results suggest the most popular summarization evaluation metric agrees with human judgments of system quality in realistic scenarios only slightly better than or equal to random chance.

This result also offers an explanation for why a naive metric such as ROUGE achieves moderately strong correlations under the standard definition of the system-level correlation (0.45 and 0.73 on \summeval{} and \realsumm{}) despite well known flaws and criticisms \citep[][among others]{PNMS05,ConroyDa08,DeutschRo20}:
it has benefited from an easy evaluation protocol.
Despite its simplicity, it is not too surprising that a large gap of 5-10 ROUGE points actually does correctly rank system pairs.
Most of its positive correlation comes from such easy examples.

\subsection{Conclusions \& Recommendations}
If it is assumed that we have enough high-quality judgments to accurately discriminate between two similarly performing systems, then the results in Fig.~\ref{fig:delta_plots} show that the correlations in realistic settings are trending very low, meaning automatic metrics are not nearly sensitive enough to distinguish between systems with only minor differences in quality.
This is problematic because this is the scenario in which metrics are most frequently used, and therefore they are not very reliable methods of evaluating summarization systems.
However, it is not all bad news.
Because the standard system-level $\tau$ values are moderately positive, consistent improvements in automatic metrics over time will likely result in better quality systems.
Similarly to stochastic gradient descent, not every reported improvement is real, but on average over time, the quality does improve.
Nonetheless, future work should focus on improving the quality of evaluation metrics when the differences in system performance are small, and researchers who compare systems should invest more effort into their human evaluations since automatic evaluations are not very reliable.

However, because the available number of system pairs to calculate the correlations in Fig.~\ref{fig:delta_plots} is rather small --- especially when evaluating on the closest system pairs --- and recent work suggests we may not have enough human judgments to accurately distinguish between similarly performing systems \citep{WeiJi21}, it could be difficult to reach any definitive conclusions about the metrics' correlations.
That being said, these are our best estimates of the correlations with the available data.
Not knowing how much we can trust automatic metrics is not a good outcome.
In this scenario, future work should focus on collecting more, high-quality human judgments so that we can better meta-evaluate automatic metrics.
Since we argue that it is important to distinguish between similarly performing systems, new data collection efforts should consider using targeted pairwise judgments between those systems instead of direct assessments across a variety of systems of diverse quality.

We recommend that proposals of new evaluation metrics also report correlations on system pairs with various differences in scores in addition to the standard system-level correlation definition.
Reporting this information would better inform users of metrics about how likely humans would agree their observed improvement is real based on its value.

\section{Related Work}
The methodology behind meta-evaluating summarization evaluation metrics was established during the DUC/TAC shared tasks \citep[][among others]{DangOw08}.
In addition to competitions for developing high-quality summarization systems, there were also shared tasks for creating automatic metrics that correlated well with human judgments.
The benchmark datasets created during DUC/TAC were small in size by today's standards because they were manually collected multi-document summarization datasets, which are hard to create at scale.
As such, all of the model-generated summaries on the full test set were judged (so $\mjudged = \mtest$; \S\ref{sec:corr}), unlike for current datasets which are too large to fully judge.

Recently, there has been growing interest in revisiting the meta-evaluation of automatic evaluation metrics for summarization, in part due to the large differences between currently popular summarization datasets and those used in DUC/TAC.
We view our work as continuing this direction of research, described next.

\citet{Peyrard19} argues that current evaluation metrics do not work as well when they are used to evaluate high-performing systems compared to those which were evaluated in DUC/TAC.

Both \citet{FKMSR21} and \citet{BGALN20} re-evaluated how well existing evaluation metrics work on the popular CNN/DailyMail dataset \citep{NallapatiBoCiCaBi16} by collecting judgments of summary quality using recent state-of-the-art systems.
These datasets were used in our analyses.
While the goal of these works was to identify which metrics correlated best with human judgments, our goal is to point out the ways in which the current methodology of meta-evaluating metrics is inconsistent with how they are used.

Then, the work of \citet{DeutschDrRo21} proposed statistical methods for estimating and comparing correlation values.
In contrast to our work, they provide statistical tools for analyzing correlations, whereas we propose new definitions of correlations.

Finally, \citet{WeiJi21} provided a theoretical analysis of the bias and variance of automatic and human evaluations of machine translations and summaries.
Among their conclusions, they argue for evaluating metrics with pairwise accuracy (Kendall's $\tau$) and that it may be prohibitively expensive to collect enough human judgments to distinguish between two systems with very similar quality.
Our work further argues that metrics should be evaluated with a variant of Kendall's $\tau$ calculated using realistic system pairs (\S\ref{sec:delta_correlations}).
Unfortunately, their results suggest that collecting enough human judgments to accurately measure how well automatic metrics perform in this setting may be very difficult.

Related studies to ours have examined how the choice of which systems to include in metric evaluations impacts the correlation values.
Both \citet{MathurBaCo20} and \citet{BGALN20} identify that metrics perform worse when scoring only the top-$k$ systems in machine translation and summarization, respectively, and examine the use of pairwise comparisons for metric evaluation.
Further, \citet{MathurBaCo20} demonstrate that outlier systems have an out-sized influence on the correlation values and recommend removing them from the metric evaluations.
In contrast, our work proposes to change the evaluation methodology for metrics so that it more closely resembles how they are used in practice.
This results in evaluating only on system pairs which are realistically compared by researchers, that is, those separated by small margins in automatic metric scores.
We believe that this is a more principled approach to how to select which system pairs to evaluate on compared to previous work.

\section{Conclusion}
In this work, we proposed two independent changes to how the system-level correlation of metrics is calculated to better align with how they are used to evaluate systems.
Our analyses showed that these modifications led to lower-variance estimates of correlations and that commonly reported improvements in metric scores may not reliably predict how humans would judge system quality.
The results from the analyses point to the need for future data collection efforts of high-quality human judgments and improving automatic evaluation metrics when differences in system performance are small.

\section*{Acknowledgments}
The authors would like to thank the anonymous reviewers for their insightful feedback on our work.

This work was supported by Contracts FA8750-19-2-1004 and FA8750-19-2-0201 with the US Defense Advanced Research Projects Agency (DARPA). Approved for Public Release, Distribution Unlimited. The views expressed are those of the authors and do not reflect the official policy or position of the Department of Defense or the U.S. Government.

This research is based upon work supported in part by the Oﬃce of the Director of National Intelligence (ODNI), Intelligence Advanced Research Projects Activity (IARPA), via IARPA Contract No. 2019-19051600006 under the BETTER Program. The views and conclusions contained herein are those of the authors and should not be interpreted as necessarily representing the oﬃcial policies, either expressed or implied, of ODNI, IARPA, the Department of Defense, or the U.S. Government. The U.S. Government is authorized to reproduce and distribute reprints for governmental purposes notwithstanding any copyright annotation therein.

This research is supported by a Focused Award from Google.

The second author is supported by the Eric and Wendy Schmidt Postdoctoral Award for Women in Mathematical and Computing Sciences.

\bibliography{acl}
\bibliographystyle{acl_natbib}


\appendix

\section{Additional Confidence Interval Results}
\label{sec:appendix_cis}
In addition to the \textsc{Boot-Inputs} CI method proposed by \citet{DeutschDrRo21}, the authors also proposed \textsc{Boot-Systems} and \textsc{Boot-Both}.
Each of the three methods makes assumptions about whether the set of $N$ systems and $M$ input documents are fixed or variable during the bootstrapping calculation.
For instance, \textsc{Boot-Inputs} assumes the $N$ systems are always the same and the $M$ input documents are random, then subsequently resamples $M$ input documents on each bootstrapping iteration to calculate the confidence interval.
\textsc{Boot-Systems} does the opposite by resampling which $N$ systems are used while holding the original $M$ input documents fixed.
\textsc{Boot-Both} assumes both the systems and inputs are variable.

\begin{figure}[t]
    \centering
    \includegraphics[width=\columnwidth]{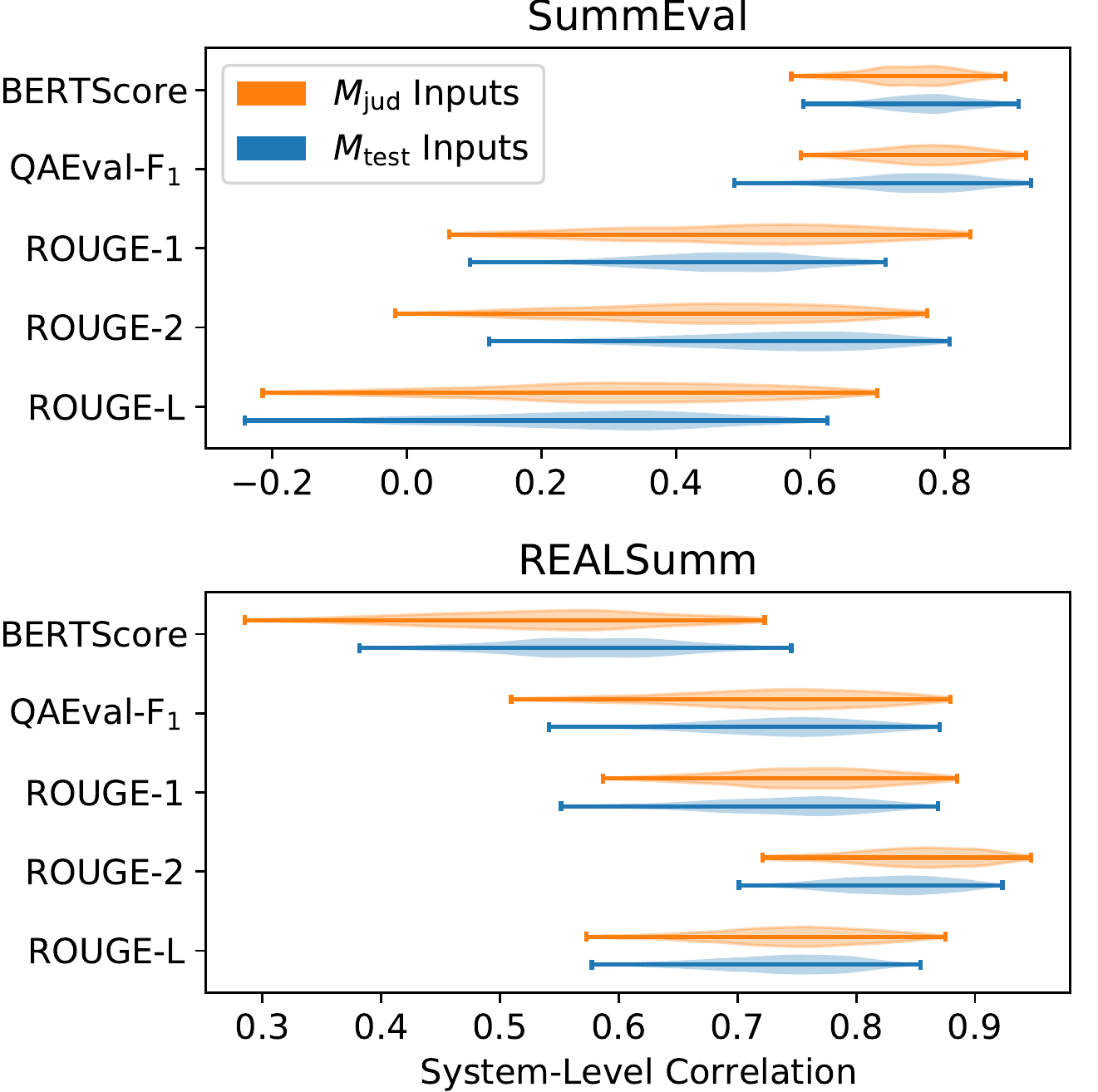}
    \caption{The 95\% CIs calculated using the \textsc{Boot-Systems} bootstrapping method with $\mjudged$ summaries in orange and $\mtest$ in blue.}
    \label{fig:ci_boot_systems}
\end{figure}

\begin{figure}
    \centering
    \includegraphics[width=\columnwidth]{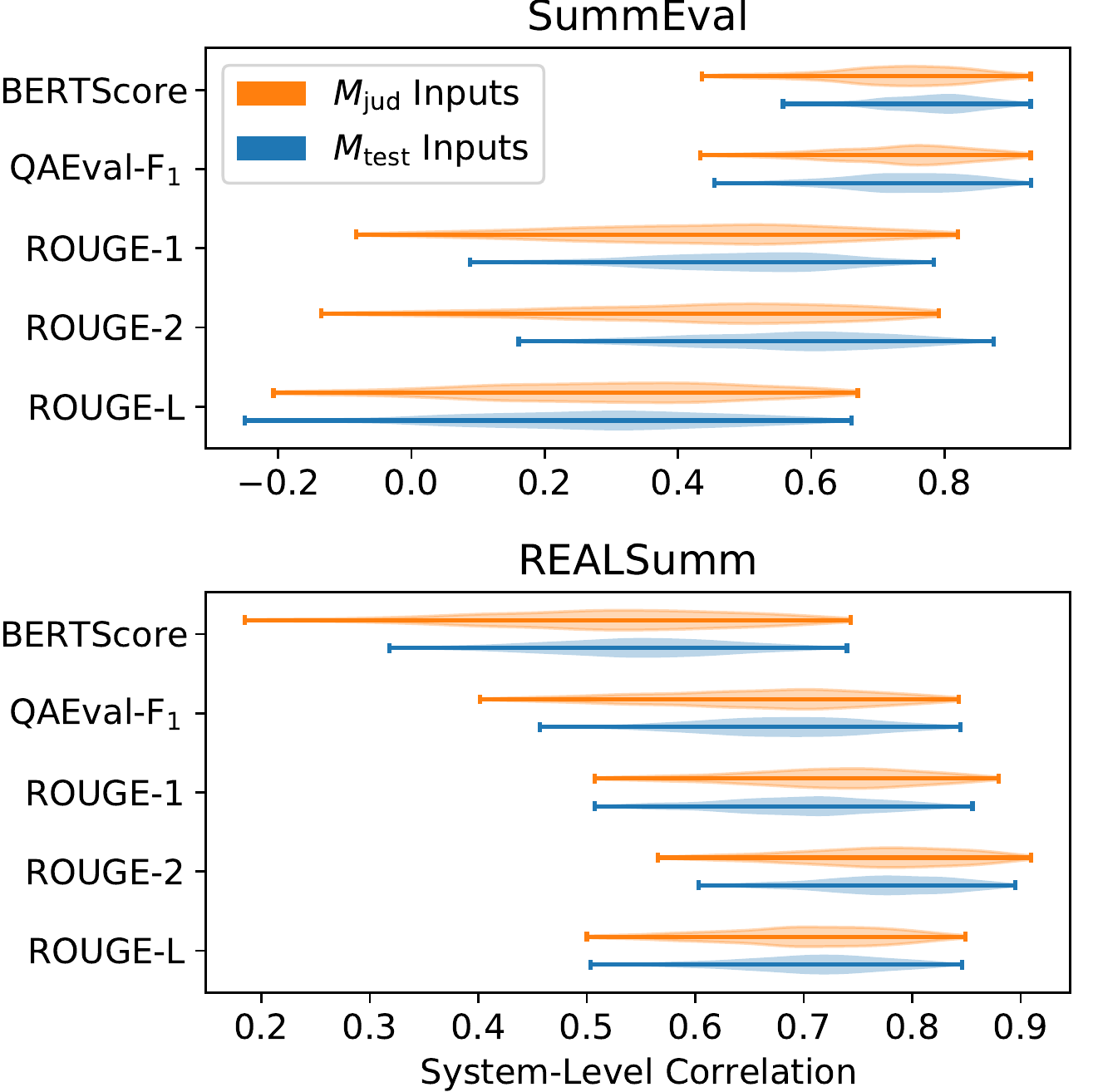}
    \caption{The 95\% CIs calculated using the \textsc{Boot-Both} bootstrapping method with $\mjudged$ summaries in orange and $\mtest$ in blue.}
    \label{fig:ci_boot_both}
\end{figure}

Figs.~\ref{fig:ci_boot_systems} and \ref{fig:ci_boot_both} contain the 95\% CIs for ROUGE, \bertscore{}, and \qaeval{} on the \summeval{} and \realsumm{} datasets using the \textsc{Boot-Systems} and \textsc{Boot-Both} methods calculated using all $\mtest$ test instances and only the $\mjudged$ annotated instances (\textsc{Boot-Inputs} included in the main body of the paper, Fig.~\ref{fig:ci_boot_input}).
The widths of the \textsc{Boot-Both} CIs decreased by 14\% and 12\%, whereas the \textsc{Boot-Systems} CIs only decreased by 1\% and 6\%.

The \textsc{Boot-Systems} widths likely decreased less because its estimation of $\rsys$ is not dependent on the variance of the system score estimates.
Since the set of $M$ input documents is fixed, the system scores do not change at all during bootstrapping, so increasing the number of summaries used to estimate those scores should not have a major effect on the estimation of $\rsys$.


\section{Additional $\rsys\Delta(\ell,u)$ Results}
\label{sec:appendix_heatmaps}
\begin{figure*}[t]
    \centering
    \includegraphics[width=\textwidth]{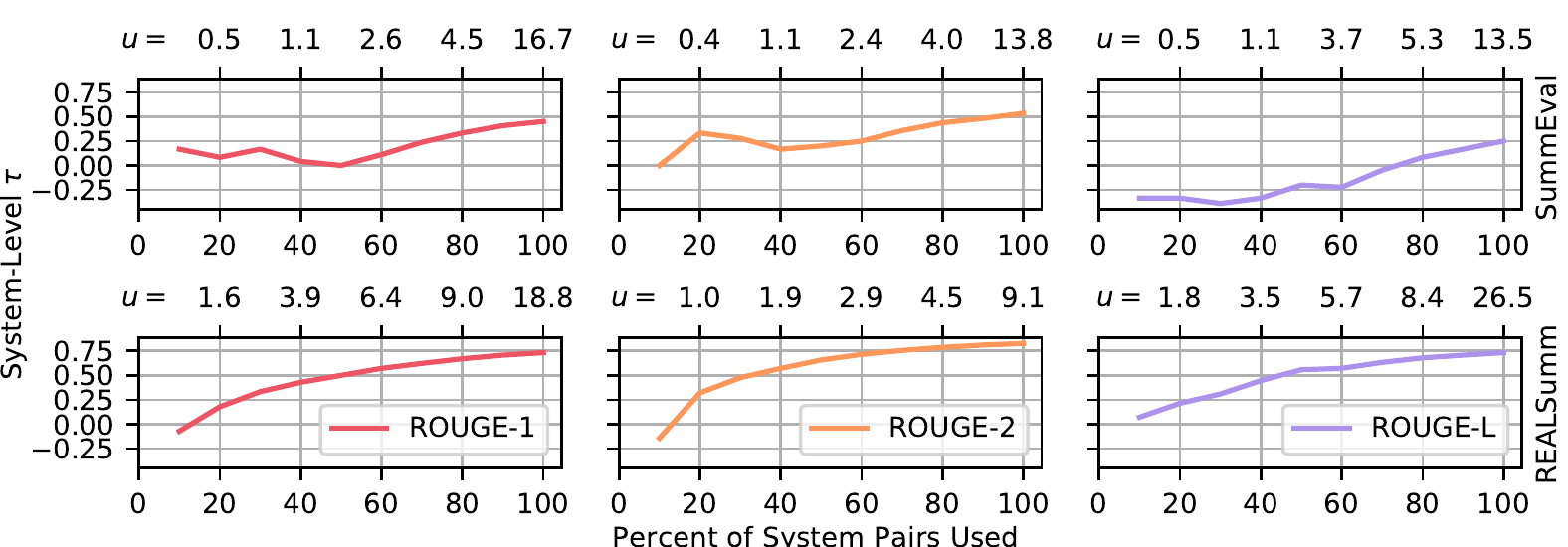}
    \caption{
    The $\rsys\Delta(\ell,u)$ correlations on the \summeval{} (top) and \realsumm{} (bottom) datasets for $\ell = 0$ and various values of $u$ for ROUGE-1, ROUGE-2, and ROUGE-L.
    The $u$ values were chosen to select the $10\%, 20\%, \dots, 100\%$ of the pairs of systems closest in score.
    Each $u$ is displayed on the top of each plot.
    }
    \label{fig:delta_plots_rouge}
\end{figure*}

Fig.~\ref{fig:delta_plots_rouge} contains the $\rsys\Delta(\ell,u)$ correlations for when $\ell=0$ for ROUGE-1, ROUGE-2, and ROUGE-L, equivalent to those shown in Fig.~\ref{fig:delta_plots} in the main body of the paper (ROUGE-1 is shown in both).
The ROUGE-2 and ROUGE-L results are largely consistent with those of ROUGE-1. The metrics' correlations to human annotations are low (or even negative) when the differences between system scores are small.
As more pairs are added that differ by larger margins, the correlations increase.

\begin{figure*}
\centering
\begin{tabular}{cc}
    \includegraphics[width=\columnwidth]{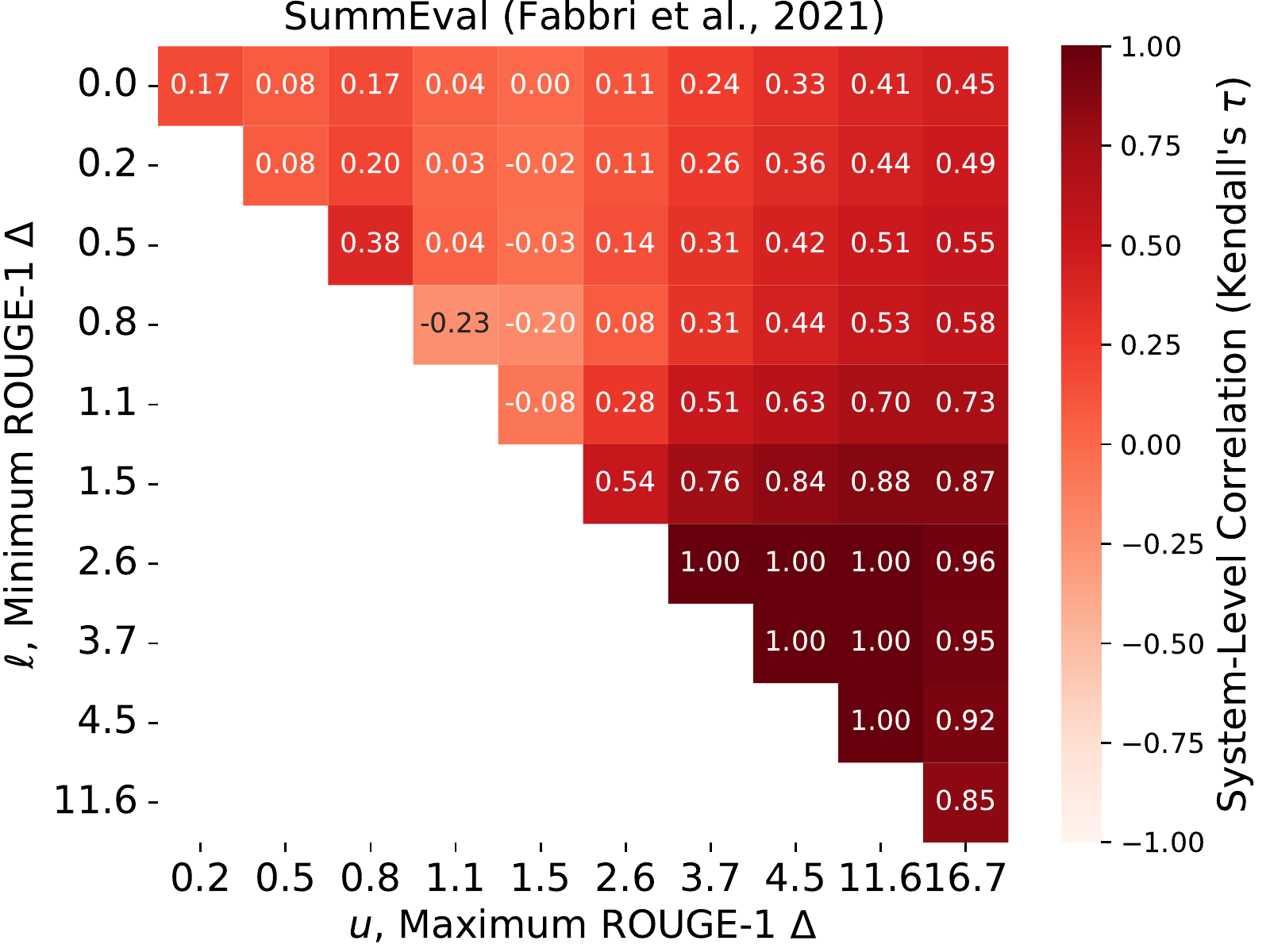} & \includegraphics[width=\columnwidth]{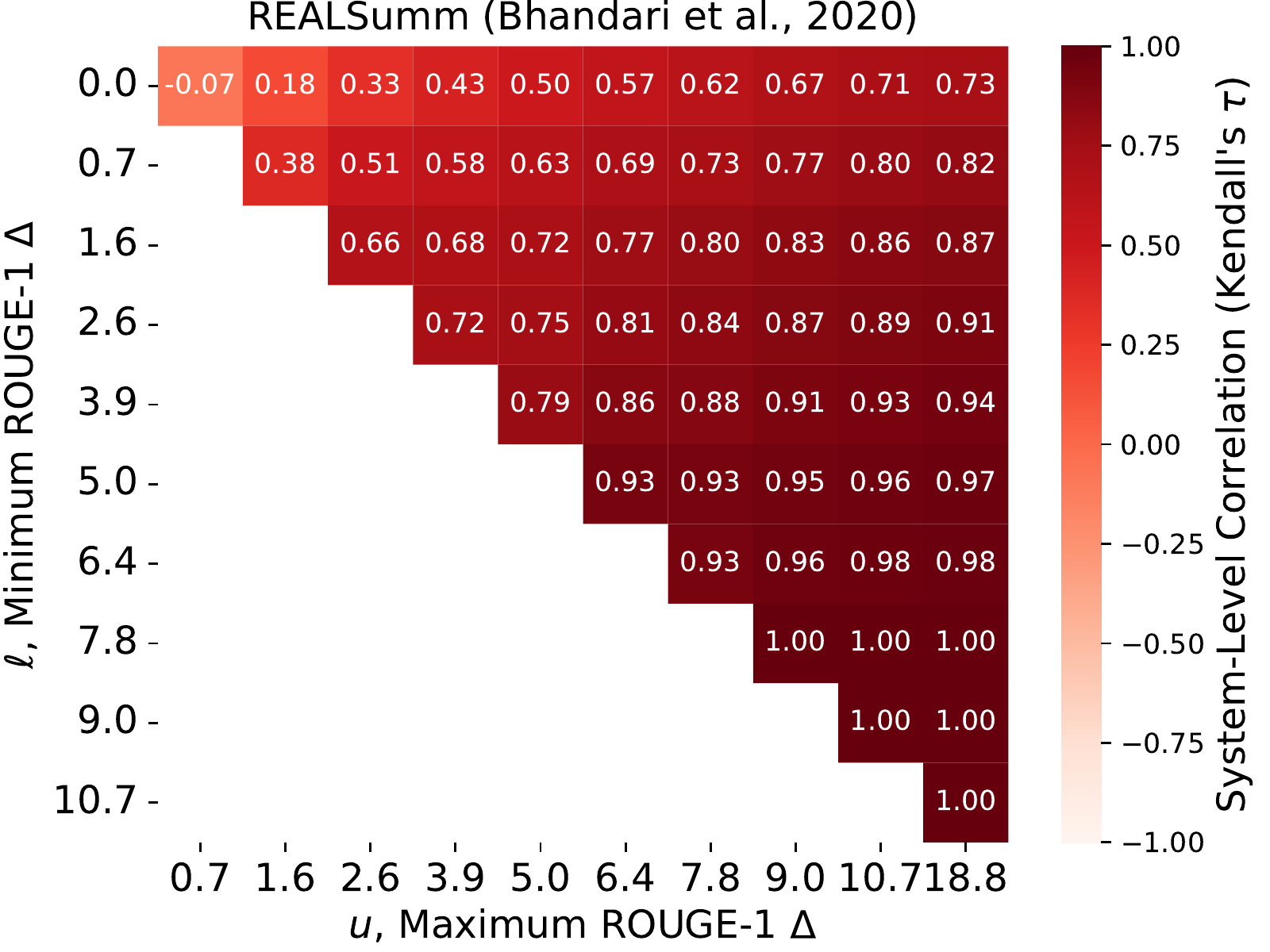} \\[1cm]
    \includegraphics[width=\columnwidth]{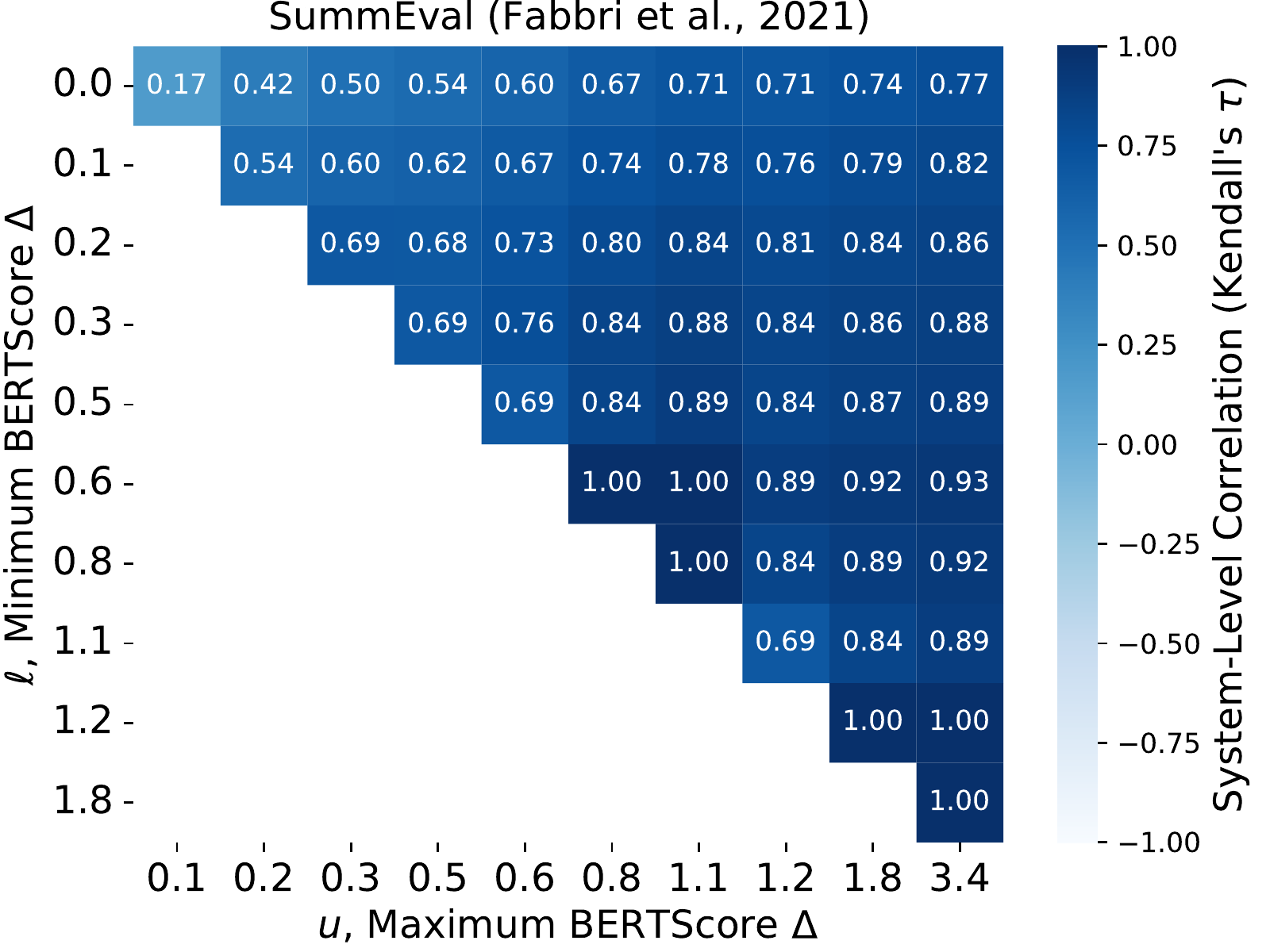} & \includegraphics[width=\columnwidth]{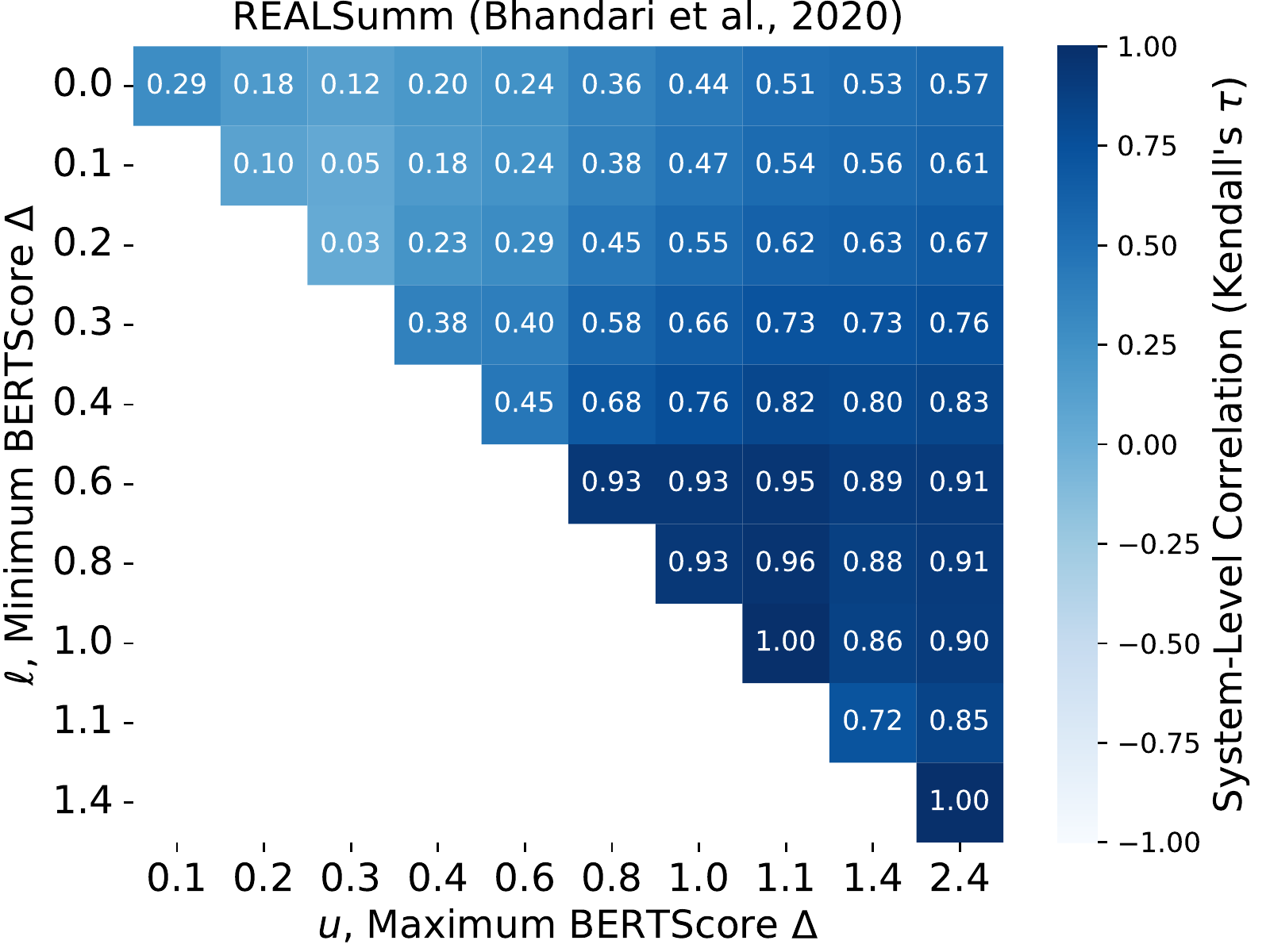} \\[1cm]
    \includegraphics[width=\columnwidth]{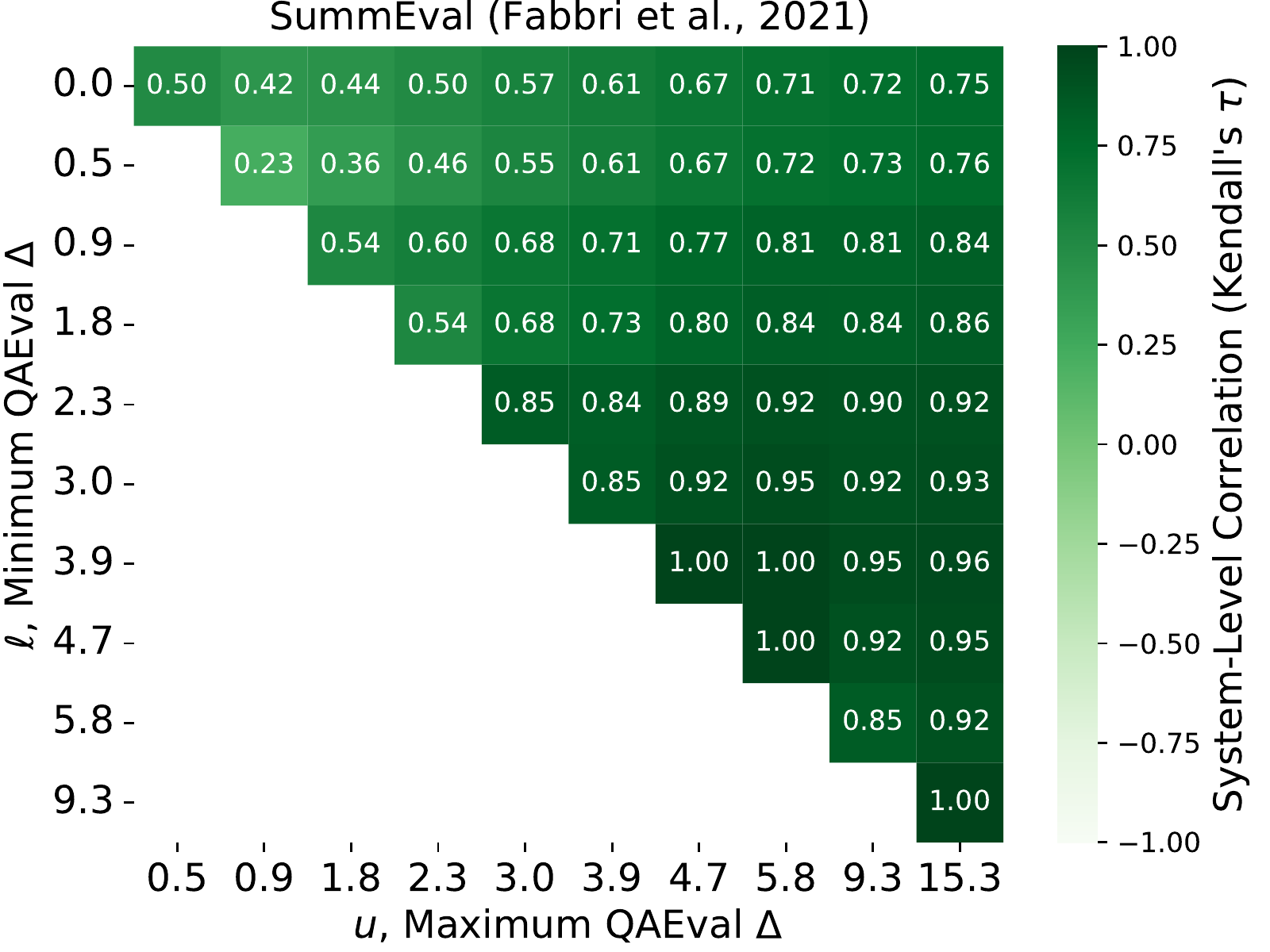} & \includegraphics[width=\columnwidth]{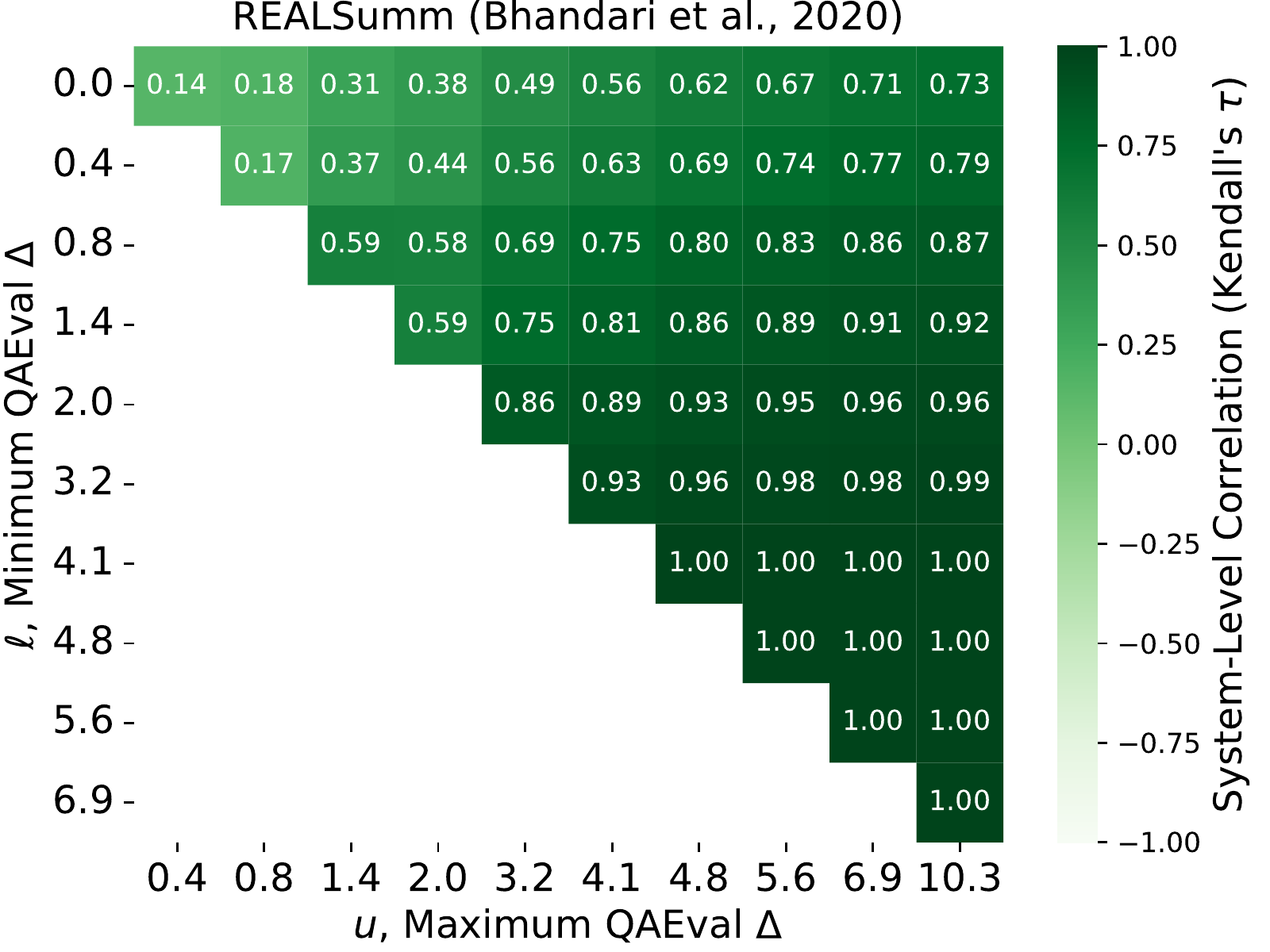} \\
\end{tabular}
\caption{
$\rsys\Delta(\ell,u)$ correlations for various combinations of $\ell$ and $u$ (see \S\ref{sec:delta_correlations_subsec}) for ROUGE (top), \bertscore{} (middle), and \qaeval{} (bottom) on \summeval{} (left) and \realsumm{} (right).
The values of $\ell$ and $u$ were chosen so that each value in the heatmaps evaluates on 10\% more system pairs than the value to its left.
For instance, the first row evaluates on $10\%, 20\%, \dots, 100\%$ of the system pairs.
The second row evaluates on $10\%, 20\%, \dots, 90\%$ of the system pairs, never including the $10\%$ of pairs which are closest in score.
The first row of each of the heatmaps is plotted in Fig.~\ref{fig:delta_plots}.
The correlations on realistic score differences between systems are in the upper left portion of the heatmaps and contain the lowest correlations overall.
Evaluating on all pairs is the top-rightmost entry, and the ``easiest'' pairs (those separated by a large score margin) are in the bottom right.
}
\label{fig:heatmaps}
\end{figure*}

\begin{figure*}
\centering
\begin{tabular}{cc}
    \includegraphics[width=\columnwidth]{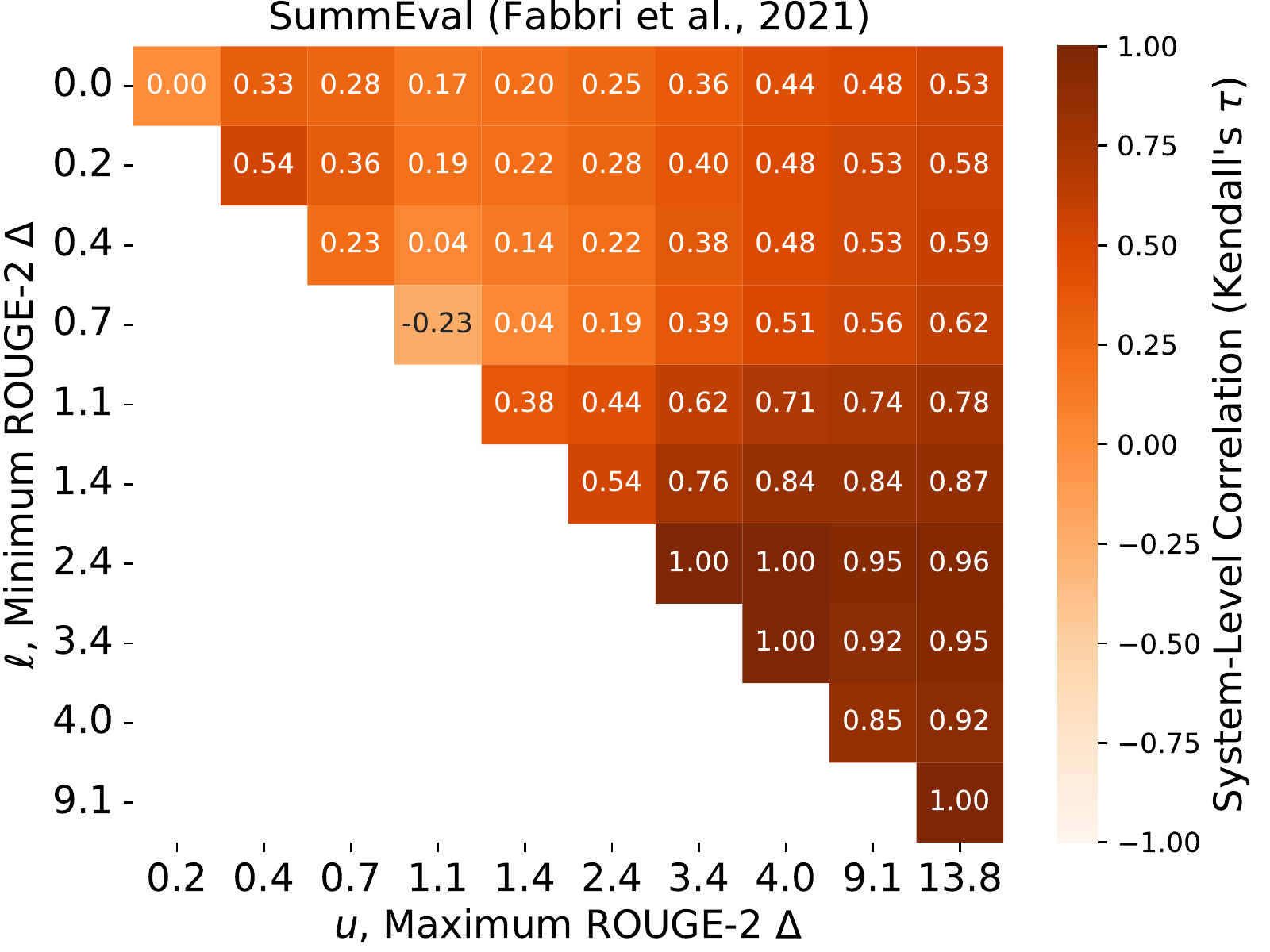} & \includegraphics[width=\columnwidth]{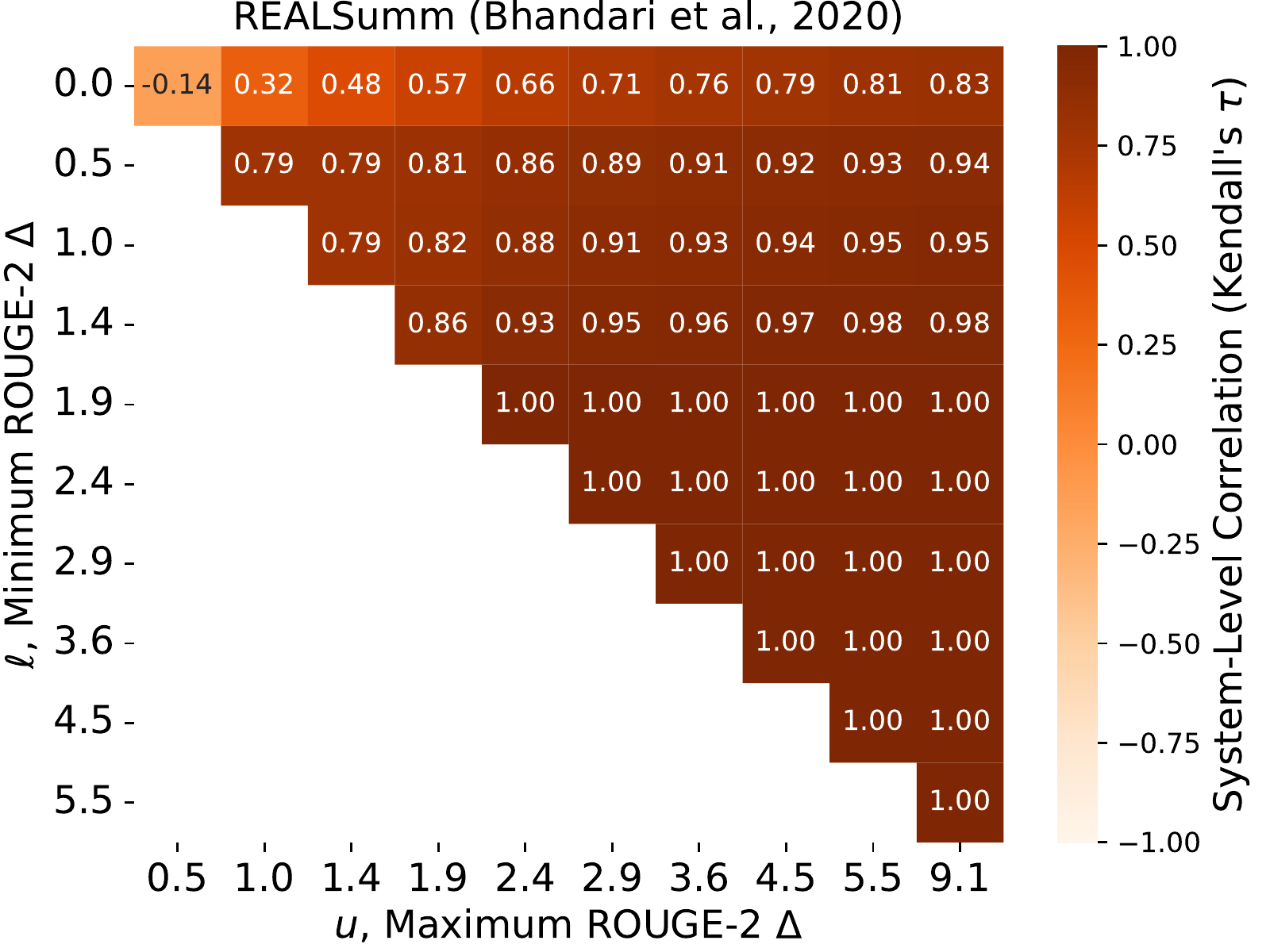} \\[1cm]
    \includegraphics[width=\columnwidth]{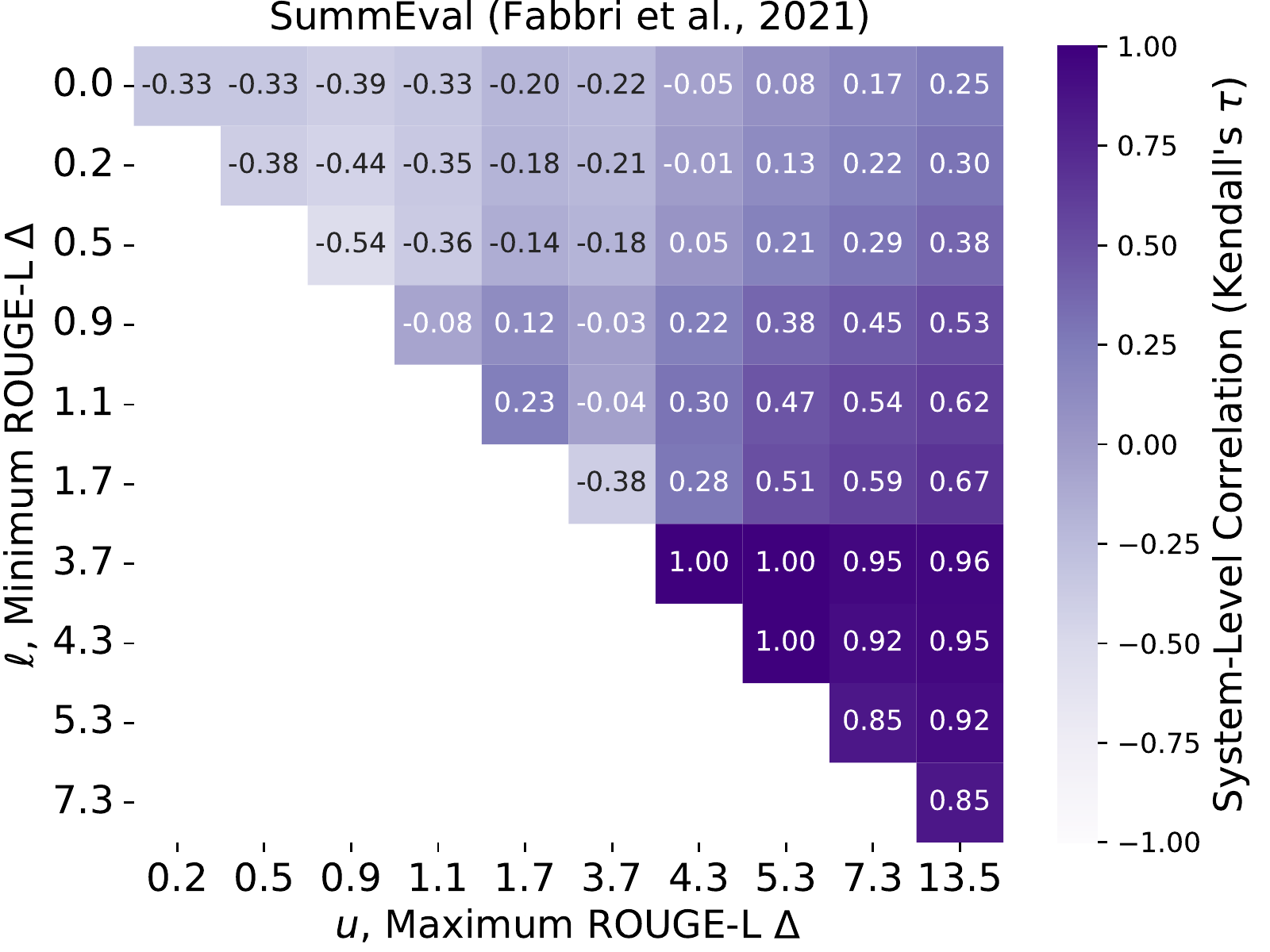} & \includegraphics[width=\columnwidth]{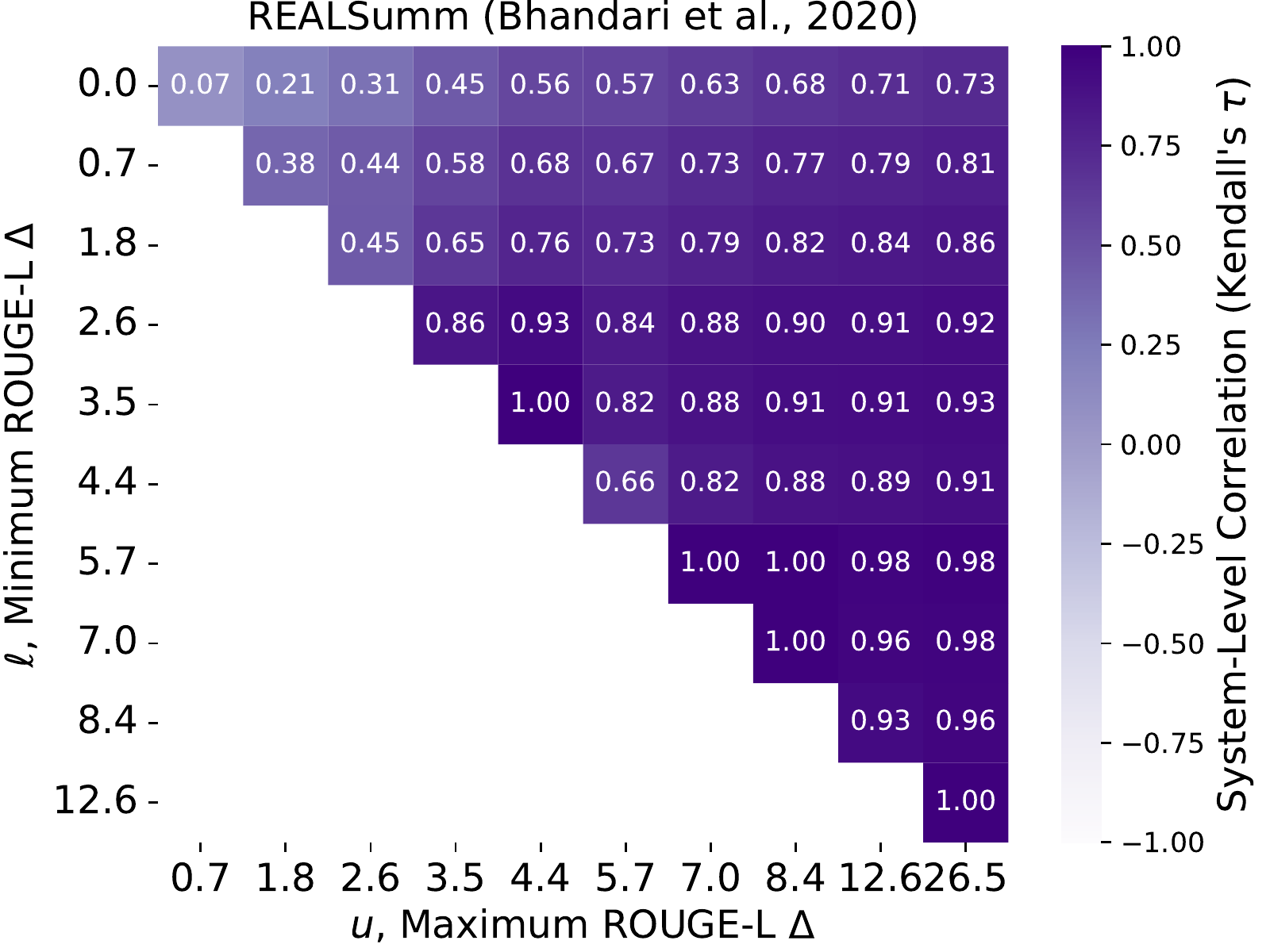} \\[1cm]
\end{tabular}
\caption{
See Fig.~\ref{fig:heatmaps} for a description of the heatmaps, shown here for ROUGE-2 (top) and ROUGE-L (bottom).
}
\label{fig:heatmaps_extra}
\end{figure*}

Figs.~\ref{fig:heatmaps} and \ref{fig:heatmaps_extra} contain the $\rsys\Delta(\ell,u)$ correlations for ROUGE, \bertscore{}, and \qaeval{} for various combinations of $\ell$ and $u$ on both the \summeval{} and \realsumm{} datasets.
The first rows of each heatmap are plotted in Figs.~\ref{fig:delta_plots} and \ref{fig:delta_plots_rouge}.

We see that as the allowed score gap between system pairs is allowed to increase (i.e., adding ``easier'' pairs to rank), the correlation increases by a large margin over the correlation on pairs close in score.
All of the metrics have nearly perfect correlation when the system pairs are separated by large margins.

\section{Summarization Paper Survey}
\label{sec:survey}

To estimate the difference in ROUGE-1 score that is commonly reported in papers, we performed a survey of recently published summarization papers.
We selected papers from 2020 and 2021 that were published in a *ACL conference (including Findings), had ``summary'' or ``summarization'' in the title, proposed a new system, and compared systems on the CNN/DailyMail dataset with ROUGE.
We selected the differences between the best two models that were compared with ROUGE on the test set.
We did not include ablation experiments for which the differences are likely smaller than the differences between the top two performing systems.
The results are shown in Table~\ref{tab:survey}.
The average reported difference was found to be 0.49 ROUGE-1.

\begin{table*}[]
    \centering
    \begin{adjustbox}{width=\textwidth}
    \begin{tabular}{llclccc}
        \toprule
        \bf Paper & \bf Model 1 & \bf Score 1 & \bf Model 2 & \bf Score 2 & $\Delta$ \\
        \midrule
        \citet{XLYWHZ20} & SAGCopy Indegree-2 & 42.56 & MASS+Copy & 41.71 & 0.85 \\
        \citet{XGCL20} & \textsc{DiscoBert w.} $\mathcal{G}_R$ \& $\mathcal{G}_C$ & 43.77 & BERTSum & 43.25 & 0.52 \\
        \citet{HuangWuWa20} & BART & 44.16 & ASGARD-DOC + $R_\textrm{ROUGE} + R_\textrm{Cloze}$ & 43.93 & 0.23 \\
        \citet{LiWuLi20} & EDUSum & 41.40 & Fast-Abs & 40.88 & 0.52 \\
        \citet{ZLCWQH20} & \textsc{MatchSum} (RoBERTa-Base) & 44.41 & BERTSum & 43.85 & 0.56 \\
        \citet{WDZWTCZ20} & BERTSum+TA & 43.06 & BERTSum & 42.13 & 0.93 \\
        \citet{JCTFCW20} & HAHSum-Large & 44.68 & MatchSum-Base & 44.41 & 0.27 \\
        \citet{ZZLWZ20} & STEP (GIGA-CM) & 44.07 & UniLM & 43.47 & 0.60 \\
        \citet{YZGZHD20} & TED 10L8H & 38.73 & Pretrained 10L8H & 38.38 & 0.35 \\
        \citet{DesaiXuDu20} & MatchSum + CUPS & 44.69 & MatchSum & 44.41 & 0.28 \\
        \citet{XWHJ20} & \textsc{Imp} + BERT (MLM) & 37.53 & \textsc{Imp} + XLNet (PLM) & 37.04 & 0.49 \\
        \citet{JinWa20} & Ours & 41.70 & CopyTransformer & 41.39 & 0.31 \\
        \citet{JCFZFLW21} & DifferSum-Large & 44.70 & MatchSum-Base & 44.41 & 0.29 \\
        \citet{NSZNMNZWAX21} & Q-C-O & 44.70 & MLE & 44.24 & 0.46 \\
        \citet{XingXiCa21} & Shuffling & 41.00 & Our Method & 40.88 & 0.12 \\
        \citet{LiuLi21} & SimCLS & 46.67 & GSum & 45.94 & 0.73 \\
        \citet{PadmakumarHe21} & PacSum & 40.26 & Lead-k & 39.69 & 0.57 \\
        \citet{BrPLRCT21} & Explicit-Structure Attention & 39.63 & Pointer-Generator + Coverage & 39.07  & 0.56 \\
        \citet{HuangKu21} & DiscoBERT & 43.77 & Proposed & 43.61 & 0.16 \\
        \citet{LiuShLa21} & UniLMv2 + SKD + Noisy T + Noisy S & 43.77 & UniLMv2 & 43.45 & 0.43 \\
        \citet{ChenYa21} & S-BART w. Discourse \& Action & 46.07 & Multi-View Seq2Seq & 45.56 & 0.51 \\
        \citet{LiuDoLi21} & GSum-Fine-Tuned & 46.18 & Base & 45.93 & 0.25 \\
        \citet{DLHJN21} & BART + MatchSum & 45.94 & BART & 44.66 & 1.28 \\
        \midrule
        Average & & & & & 0.49 \\
        \bottomrule
    \end{tabular}
    \end{adjustbox}
    \caption{A survey of recent summarization papers published in *ACL conferences and the differences in ROUGE-1 score they reported on the CNN/DailyMail dataset.
    \citet{NMAPBM20} is not included because it is not clear which results were considered comparable to their model.}
    \label{tab:survey}
\end{table*}

\end{document}